\documentclass{article}

\usepackage{microtype}
\usepackage{graphicx}
\usepackage{subfigure}
\usepackage{booktabs} % for professional tables
\usepackage{multirow}

\usepackage{hyperref}

\usepackage[accepted]{icml2024}

\usepackage{amsmath}
\usepackage{amssymb}
\usepackage{mathtools}
\usepackage{amsthm}

\usepackage[capitalize,noabbrev]{cleveref}

\theoremstyle{plain}

\theoremstyle{definition}

\theoremstyle{remark}

\usepackage[textsize=tiny]{todonotes}

\icmltitlerunning{Self-Supervised Coarsening of Unstructured Grid with Automatic Differentiation}

\begin{document}

\twocolumn[
\icmltitle{Self-Supervised Coarsening of Unstructured Grid with Automatic Differentiation}



\icmlsetsymbol{equal}{*}

\begin{icmlauthorlist}
\icmlauthor{Sergei Shumilin}{sk}
\icmlauthor{Alexander Ryabov}{sk}
\icmlauthor{Nikolay Yavich}{sk}
\icmlauthor{Evgeny Burnaev}{sk,airi}
\icmlauthor{Vladimir Vanovskiy}{sk}
\end{icmlauthorlist}

\icmlaffiliation{sk}{Applied AI Center, Skolkovo Institute of Science and Technology, Moscow, Russia}
\icmlaffiliation{airi}{AIRI Institute, Moscow, Russia}

\icmlcorrespondingauthor{Sergei Shumilin}{s.shumilin@skoltech.ru}
\icmlcorrespondingauthor{Alexander Ryabov}{a.ryabov@skoltech.ru}
\icmlcorrespondingauthor{Nikolay Yavich}{n.yavich@skoltech.ru}
\icmlcorrespondingauthor{Evgeny Burnaev}{e.burnaev@skoltech.ru}
\icmlcorrespondingauthor{Vladimir Vanovskiy}{v.vanovskiy@skoltech.ru}

% You may provide ay keywords that you
\icmlkeywords{differentiable physics, physics informed machine learning}

\vskip 0.3in
]



\printAffiliationsAndNotice{}  % leave blank if no need to mention equal contribution

\begin{abstract}
Due to the high computational load of modern numerical simulation, there is a demand for approaches that would reduce the size of discrete problems while keeping the accuracy reasonable. 
In this work, we present an original algorithm to coarsen an unstructured grid based on the concepts of differentiable physics. 
We achieve this by employing $k$-means clustering, autodifferentiation and stochastic minimization algorithms. 
We demonstrate performance of the designed algorithm on two PDEs: a linear parabolic equation which governs slightly compressible fluid flow in porous media and the wave equation.
Our results show that in the considered scenarios, we reduced the number of grid points up to 10 times while preserving the modeled variable dynamics in the points of interest.
The proposed approach can be applied to the simulation of an arbitrary system described by evolutionary partial differential equations.  
\end{abstract}

\section{Introduction}
\label{submission}

%Why this is important.
Modelling of fluid flow in general and subsurface fluid flow in particular
are known to be highly computationally demanding problems. However, such
problems arise in various practical engineering applications.
High computational demand comes from the need to 
%honor 
deal with detailed time and spatial resolutions as well as possible non-linearity of
governing partial differential equations. On the practical side,
the problem is even worse since
modellers typically run simulations several times 
with different sets of parameters.

% convential upscaling
There are different approaches that tend to reduce the computational load
of these problems. 
In the field of subsurface fluid flow, one of the most traditional approach is \textit{upscaling}, \cite{Qi_Hesketh_2005}.
The goal is to substitute the original fine grid with a coarser grid and to upscale the equation coefficients (permeability, porosity and other), i.e. to transfer fine grid values 
to cells on the coarse grid.

In other words, upscaling is a means to 
substitute a heterogeneous region with a homogeneous. 
The standard approaches use arithmetic or harmonic averages over coarse grid blocks (clusters).
There are also more advanced approaches that estimate upscaled coefficients by solving the flow equations within coarse grid blocks.
The techniques differ in the way they average fine-grid data
and form coarse grid blocks
\color{black}
but 
they all tend to reduce modelling grid size and thus save computational time. 

% ROM
The main feature of the upscaling approaches 
is that they try to coarsen the problem based on the fine grid and
equation coefficients but not use the modelled data.
\textit{Reduced-order modelling} (ROM) methods work differently.
They use some modelled data (e.g. early times) to reduce computational load of the rest of the problem (late times).
The family of ROM methods was actively developing in the last decades.
One of its most well-known representatives
is proper orthogonal decomposition (POD), see e.g. \cite{kerschen2005method,rowley2005model,Guo2019} and references therein.
It approximates complex fluid flow by using several dominant spatial eigenvectors. 
Thus the method is directly related to the principle component analysis (PCA).
The dynamic mode decomposition (DMD), \cite{kutz2016book},  is another popular approach
in the recent years. It uses both spatial and temporal frequencies to represent 
fluid dynamics and predict its future state. The DMD is an equation-free 
technique, i.e. it provides an explicit solutions without
solving the reduced model.

If thinking about upscaling from the supervised learning prospective, 
the following naive idea might come up: what if we try to search the coarse grid by trying to fit physical fields modelled on coarse grid to original
grid data? However, this is not practical since data modelled with conventional 
numerical modelling algorithms is not differentiable with respect to grid nodes
locations. To address issue, we follow the concept of \textit{differentiable
physics}.   

A set of techniques known as differentiable physics simulation utilizes gradient-based methods for learning and controlling physical systems \cite{liang2020differentiable}.
The main objective is to integrate every step of the simulation in a way that allows autodifferentiation (AD) \cite{holl2020learning}.
Then we are able to compute the partial derivatives of the output with respect to the input.
AD approach is used in molecular dynamics even for spatial optimization of particle positions \cite{schoenholz2020jax}.
Also AD can be used for optimization of material structure \cite{dold2023differentiable}.
Since any PDE solver assumes system's dynamics over a time period where any current state depends on the previous states we get a very nested computational graph and it becomes one of the main challenges.

There are works that use self-supervised metric in order to modify the computational mesh. For example, \cite{FIDKOWSKI2021109957} leverage the power of neural networks to optimize mesh anisotropy based on error indicators and Hessian matrices from primal and adjoint solutions. 
However, our method is different in several aspects.
It extends the concept of mesh optimization by incorporating differentiable physics into the coarsening process.
This allows for the end-to-end gradient-based optimization of the mesh, ensuring that refinements are intrinsically aligned with the underlying physical laws. 
Our method utilizes the solution at the measurement points only and does not require the Hessian, which is tedious to evaluate and store. 
Additionally, employing self-supervised learning through differentiable Voronoi tessellation and the finite volume solver, our method achieves grid coarsening without using training data from other mesh adaptation methods as in \cite{FIDKOWSKI2021109957}.
\cite{park2016unstructured} provides an overview of traditional methods of adaptation of structured meshes. 
However, all of the comparisons described above differ our method from the library of methods described in \cite{park2016unstructured}.
% GNN?
In this work, we present an original approach to reduce computational modelling 
complexity. % that combines some ideas of the mentioned above approaches.
Specifically, we design a coarse grid by iterative minimization of the 
misfit between data modelled on coarse and fine grids.
We achieve this by employing $k$-means clustering, autodifferentiation
and stochastic minimization algorithms. As far as the authors are concerned, 
this is the first attempt to design an unstructured grid coarsening algorithm
based on autodifferentiation.
We look at slightly compressible fluid flow in porous media,
which is governed by a linear parabolic equation, yet the idea is applicable to an arbitrary evolutionary partial differential equation (we tested the approach also on a hyperbolic wave equation).

Our main \textbf{contributions} are as follows:
\begin{enumerate}

\item \label{c1} We make use of graph representation of finite volume solvers that allows to model PDEs in a differentiable way by widely used in GNNs Message Passing technique;
\item \label{c2} We propose a way to make simulation fully differentiable by combining differentiable Voronoi tessellation with the finite volume solver introduced at \ref{c1};
\item We propose a pipeline allowing to coarsen the original unstructured grid whilst preserving the quality of simulation. Such a result is achieved by optimizing the difference between simulation results on two grids using the differentiability of simulation introduced at \ref{c2}.

The code is available on GitHub\footnote{\tiny https://github.com/SergeiShumilin/DifferentiableUnstructuredGridCoarsening}.

\end{enumerate}

\begin{table}[t]
\caption{Main notations and acronyms}
\label{tab:notations}
\vskip 0.15in
\begin{center}
\begin{small}
\begin{sc}
\begin{tabular}{ll}
\toprule
\textbf{Notion} & \textbf{ Notation}  \\
\midrule
    modelling domain & $V$\\
    domain boundary  & $B$ \\
    original grid size & $N$ \\
    Voronoi sites/point cloud& $S = \{S_i\}_{i=1}^N$,  \\ 
    & \quad  $S_i\in V$ \\
    Voronoi cells/regions & $V_i$, $1\leq i \leq N$\\
    VT of $S$ & $V(S) =\{V_i\}_{i=1}^N$ \\
    finite volume method & FV or FVM \\
    time interval & $0<t<T$\\ 
    number of time steps&$m$\\
    time step & $\tau = T/m$ \\ 
    pressure& $p(x,y,t)$\\ 
    source term & $f(x,y,t)$ \\ 
%         pressure at $t=0$ & $p^0(x,y)$ \\ 
    permeability&$K(x, y)$\\ 
    mean permeability within cell $i$ & $K_i$ \\ 
    mean source within & $f_i^k$ \\ 
    \quad cell $i$ at time step $k$ &  \\ 
%    FV pressure within & \\
%    \quad cell $i$ at time step $k$ & $p_i^k$ \\
    FV pressure at time step $k$    & $p^k \in \mathbb{R}^N$ \\ 
    FV pressure time series in cell $i$ & $p_i \in \mathbb{R}^m$\\
%          pressure at $i$-th point [Pa]&$p_i(t)$\\
 %         
%         $Card(\mathbf{S})$ & $N$\\
%
% tuple of $S$ and $K$ &$P=(S, K(x, y))$\\
    coarse grid size & $n$ \\
    degree of reduction & $r=n/N$ \\
    aggregation function & $\Theta$ \\
\bottomrule
\end{tabular}
\end{sc}
\end{small}
\end{center}
\vskip -0.1in
\end{table}

\begin{figure*}[hbt!]
\begin{center}
\includegraphics[width=\textwidth]{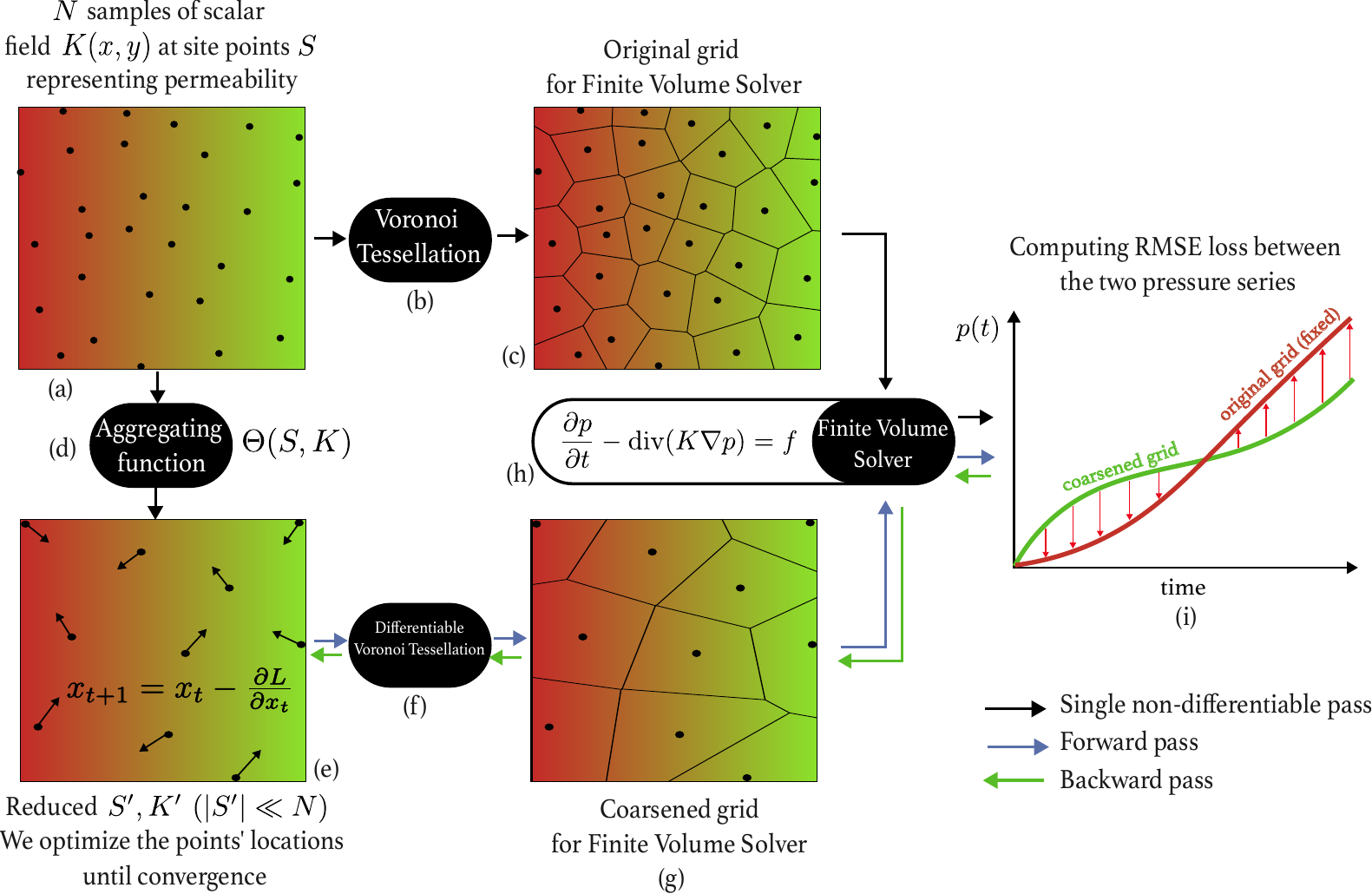}
\caption{The general pipeline of the method. The main idea is to reduce the number of points in the original field but preserve the quality of simulation. The method does this in a self-supervised manner. \textbf{a)} As an input the method obtains the 2D point cloud, permeabilities in the points and boundary $B$ ; \textbf{b)} the input point cloud is processed into the Voronoi tessellation; \textbf{c)} geometric parameters of the tessellation including the lengths of Voronoi edges and the areas of the Voronoi regions which further are processed into the differentiable graph FV solver; \textbf{d)} Aggregating function which reduced the number of points; \textbf{e)} the locations of the reduced points are iteratively optimized in order to reduce the loss; \textbf{f)} Differentiable Voronoi tessellation; \textbf{g)} coarsened grid where all geometric parameters represented as tensors and are included  into computational graph; \textbf{h)} Finite volume solver which operates on tensors and implements explicit Euler scheme; \textbf{i)} loss function that compares the $p(t)$ at some point.}
\label{teaser}
\end{center}
\end{figure*}

\section{Background}

\subsection{Autodifferentiation}

Automatic differentiation (AD) is a technique for obtaining the gradients of a program's output with respect to input \cite{Naumann}.
One of the most important algorithms based on AD is backpropagation.
Nowadays, it's primary application is in training deep learning models.
Backpropagation is proved to be powerful enough to allow training even Large Language Models with billions of parameters.

However, AD is used in much broader sense in numerous applications in physics, chemistry and biology.
A program or a neural network can be represented as a computational graph in which nodes represent functions, and edges represent input/output relationships.
Each node of the computational graph corresponds to a simple operation and, in fact, represents a function.
Simple functions then combined to form a much complex models.
The computations graph contains nodes of three types: leaf nodes which are constants or input variables, terminal nodes which represent output, and regular nodes which are none of the others. 

There is a number of Python based frameworks for AD. For our experiments we use PyTorch \cite{paszke2017automatic}. 

\section{Method}

In this work we consider two PDEs: a linear parabolic equation which governs slightly compressible fluid flow in porous media and the wave equation.
To demonstrate the concept throughout the paper we use the parabolic equation case.
We use reservoir modelling terminology to describe subsurface flow.

\subsection{Governing equations and their discretization}

Let $V$ be a polygonal domain for the reservoir. We assume that
fluid flow is controlled by injection and production wells.
A slightly compressible fluid flow in heterogeneous porous media 
is described by a parabolic equation, \cite{Chen2007}. 
With minor simplifications, it takes the following form,
\begin{equation}
    \displaystyle\frac{\partial p}{\partial t} - \textrm{div}(K \nabla p) = f,   \quad 0<t<T,
    \label{eq:parab}
\end{equation}
where $p(x,y,t)$ is the unknown pressure, $K(x,y)$ is the permeability, $f(x,y,t)$ is the source/sink
term. 
The above equation is completed with the initial condition $p = p^0(x,y)$ at $t=0$
and zero Neumann boundary conditions on the domain boundary. 

To receive a numerical solution of \eqref{eq:parab},
we applied the finite volume method for spatial discretization \cite{Herbin2000,Kuznetsov2007}
and the forward Euler scheme for temporal discretization.
%
%Applying the forward Euler scheme, 
%we receive,
%
%\begin{equation}
%    \displaystyle\frac{p^{n+1}-p^n}{\tau} - \textrm{div}(K \nabla p^n) = f^n,   \quad n=0\cdots N-1 
%    \label{eq:euler}
%\end{equation}
%
%where $p^n(x,y)$ and $f^n(x,y)$ are the pressure and source terms at $n$th time step.
We appreciate that the forward Euler scheme has the well-known stability restriction, Appendix~\ref{sec:stability}.
Implicit schemes do not have this restriction, yet the forward Euler scheme was much easier
to integrate in our pipeline. 
%We plan to investigate implicit schemes in our future work.

Let $m$ be the number of time steps of size $\tau = T/m$ and
$S_i$, $i=1\cdots N$, be arbitrary points in $V$. 
We form a Voronoi mesh using these points, and denote as $V_i$ the respective cells, $|V_i|$ its area, as $e_{ij}$ the edge separating adjacent cells $V_i$ and $V_j$, $|e_{ij}|$ its length, $h_{ij}$ distance between Voronoi cites of two adjacent cells $V_i$ and $V_j$.

Now the discretization of \eqref{eq:parab} reads as follows,
\begin{equation}
    D \displaystyle\frac{p^{k+1}-p^k}{\tau} + A  p^k = D  f^k, \quad k=0\cdots m-1,
    \label{eq:fvm}
\end{equation}
where $p^k\in\mathbb{R}^N$ and $f^k\in\mathbb{R}^N$ are discrete pressure and source vectors, respectively,
$D$ is a diagonal matrix of areas $|V_i|$ and $A\in\mathbb{R}^{N\times N}$ is the finite-volume system matrix.
It is a sparse symmetric matrix with the following entries. If cells $V_i$ and $V_j$
are adjacent, then
\begin{equation}
   A_{ij} =  -\displaystyle\frac{|e_{ij}|}{h_{ij}} \displaystyle\frac{2 K_j K_j}{K_i + K_j}, 
  \label{eq:off_diag}
\end{equation}
otherwise off-diagonal entries are zero, and the diagonal entry is equal to the negative
sum of  off-diagonal entries.
\begin{equation}
   A_{ii} =  - \sum_{j=1, j\neq i}^N  A_{ij}.
   \label{eq:diag}
\end{equation}
Since the expressions for matrix entries are fairly simple we do not actually store the matrix,
but rather we compute the product $A p^k$ on-the-fly.

%%%%%%%%%%%%%%%%%%%%%%%%%%%%%%%%%%%%%%%%%%%%%%%%%%%%%%%%%%%%%%%%%%%%%%%%%%%
\subsection{Differentiable graph simulator}

In the context of fluid dynamics simulations, particularly for slightly compressible fluids in complex geometries, we have developed a novel differentiable graph-based simulator that incorporates the finite volume method (FVM) within a graph neural network (GNN) framework. 
Our  differentiable graph simulator is implemented based on the Message Passing paradigm in graph neural networks. 
For implementing such a simulator, we used Pytorch Geometric \cite{Fey/Lenssen/2019} framework. 
The Message Passing idea has been adopted to physics simulations, enhancing the modeling of physical processes on irregular graph-structured data.
Our solver operates on tensors.
In fact, we use Message Passing to calculate divergence in \ref{eq:parab}.

Our solver uses weights for the edges of the graph. 
These weights are crucial in determining the flow dynamics across the graph and are calculated based on the physical properties of the nodes and edges, including permeability ($K$) and the distance between nodes ($h_{ij}$). 
In our case, the weights are given by negative off-diagonal values of the FVM system matrix, see.~\eqref{eq:off_diag}:
\begin{equation}
    w_{ij} = -A_{ij},
    \label{eq:weights}
\end{equation}
since every non-zero off-diagonal matrix entry corresponds to the respective edge of the graph.

The solver works by updating the properties of the physical field of interest at each node. 
It performs an update of the values of  at the nodes using explicit Euler scheme for time integration, and in our case it is the solution of equation \eqref{eq:fvm}:
\begin{equation}
     p^{k+1} = p^k + \tau (f^k -D^{-1} A p^k)
    \label{eq:pressure_update}.
\end{equation}

\subsection{Differentiable Voronoi tessellation}
The Voronoi tessellation forms the fundamental concept in computational geometry, partitioning a given space into regions based on proximity to a set of predefined points.
For a site point $s_i$, $V_i$ consists of all points in $\mathbb{R}^2$ for which $s_i$ is the nearest neighbour site point.
The common boundary part of two Voronoi regions is called a Voronoi edge if it contains more than one point.

Numerous implementations of the Voronoi tessellation written in C++ or other languages hinder the gradient flow through the tessellation itself making it impossible to build end-to-end differentiable applications.

\begin{figure}[hbt!]
\begin{center}
\includegraphics[width=\columnwidth]{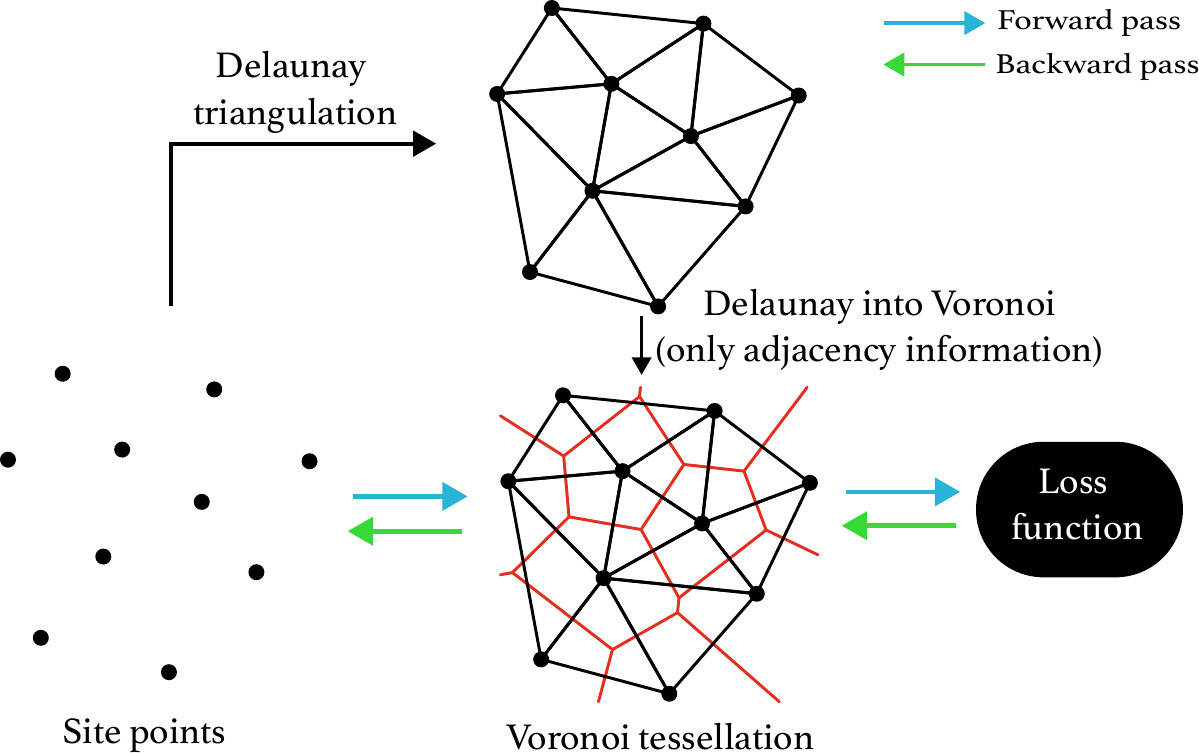}
\caption{The principle scheme of the differentiable Voronoi tessellation. The input points are Delaunay triangulated. Then the adjacency information is used for calculating the geometric parameters of the Voronoi tessellation.  Blue arrows represent the forward pass which consists of geometric calculations on tensors. Backward pass is done automatically by AD.}
\label{diff_voronoi}
\end{center}
\end{figure}

Here we use the differentiable version of the Voronoi tessellation~\cite{diffvoronoi}.
As noted before, the FV solver uses information about the partitioning of the space.
It needs both the length of the Voronoi edges and the areas of the Voronoi cells.
To make our method end-to-end differentiable, we need these geometric parameters to be the part of the computational graph.
To do so the method leverages the idea of separating the construction of the Delaunay triangulation from computing the differentiable tensors with specific information (see Fig.~\ref{diff_voronoi}).
Delaunay triangulation only provides indexing information for the construction of the Voronoi tessellation.
Then, based on the adjacency information the geometric parameters of the Voronoi tessellation are calculated.
These parameters include: lengths of the Voronoi cells edges ($e_{ij}$), areas of the Voronoi cells ($|V_{i}|$).
This is possible because of the duality between Delaunay triangulation and Voronoi tessellation \cite{Aurenhammer}.

Let $t_{a,b,c}$ be a triangle in Delaunay triangulation of $S$ supported by points $a, b, c$. 
Let $x$ and $y$ be the center of the circumcircle of $t_{a,b,c}$.
The analytical expression to calculate the circumcenter (center of circumcircle) of $t_{a,b,c}$:
\begin{multline}
x = \frac{\left(a_1^2-c_1^2+a_2^2-c_2^2\right)\left(b_2-c_2\right)}{D} - \\ 
    -  \frac{\left(b_1^2-c_1^2+b_2^2-c_2^2\right)\left(a_2-c_2\right)}{D} \label{circumcenters_x} \\
\end{multline}
\begin{multline}
y = \frac{\left(b_1^2-c_1^2+b_2^2-c_2^2\right)\left(a_1-c_1\right)}{D} - \\ 
    -  \frac{\left(a_1^2-c_1^2+a_2^2-c_2^2\right)\left(b_1-c_1\right)}{D} \label{circumcenters_y} \\
\end{multline}
where $D$ which is four times the area of $t_{a,b,c}$ , is given by
\begin{align}
D=2\left[\left(a_1-c_1\right)\left(b_2-c_2\right)-\left(b_1-c_1\right)\left(a_2-c_2\right)\right]
\end{align}
It's clear that we can calculate $\displaystyle  \frac{\partial x}{\partial a}$, $\displaystyle  \frac{\partial y}{\partial a}$ and the same for $b$ and $c$.

The lengths of the Voronoi edges are calculated based on the circumcenters of the Delaunay triangles.
The areas of the Voronoi regions are calculated by the differentiable version of the Shoelace formula \cite{braden1986surveyor}. 
Voronoi regions which overlap the boundary $B$ are clipped in a differentiable manner.

As FV solvers use the Voronoi partitioning of the space, now we are able to pass gradients from the solver to the input data.

\subsection{General pipeline}

The general pipeline for the method is presented in Fig.\ref{teaser}.
The main motivation is to obtain the coarsened version of the initial physical field that well approximates the initial physical qualities but has a much smaller size.
The main idea is to iteratively optimize the locations of points of the coarsened grid in order minimize misfit between data modelled on the coarse and fine grids, respectively.

\textbf{Input data} As an input we have a 2D point cloud which we call site points $S$ 
and values of discrete permeability field denoted as $K_S = \{K_i\}_{i=1}^M$.
Initially $K_i$ represents permeability at some site point.
After the Voronoi tessellation of $S$, $K_i$ represents mean permeability within the respective Voronoi cell, $V_i$.
Also as an input we pass the boundary $B$. %We assume $S \subset V$.

\textbf{Pooling function.}
Despite the position of site points may be random before we start optimization, we choose a good initial approximation by using a pooling function.
Moreover, we use pooling for $K_S$ as well.

Let $\Theta$ be an aggregating function that takes $S$ and $K_S$ as an input and produces $\Hat{S}$ and $\Hat{K}$ with fewer number of points.
$\hat{S}$ and $\hat{K}$ taken together, represent a 3D point cloud.
$\Theta$ is a two step function.
First, $\Theta$ performs clusterization of the site points and permeabilities into $n$ clusters.
Then it applies arithmetic mean function for both coordinates and permeabilities inside a cluster.
Thus the pooling function $\Theta$ acts as:
\begin{equation}
    \Hat{S}, \Hat{K} = \Theta(S, K_S)
    \label{psi}
\end{equation}
When apply pooling we leave source and measurement points untouched. We apply particular configuration of the pooling function but we assume that any function that satisfies the desired properties may be applied.

\textbf{Theoretical formulation of the method.} 
Let $p_{s}(S)$ and $p_{s}(S^*)$ be the time series for modelled variable (pressure in our case) in a measurement point $s$, $s \in S$ and $s \in S^{*}$ because we keep it untouched.
We assume that both time series are of equal length.
Theoretically we formulate our procedure as a function $OPTIMIZE$.
It outputs optimized coordinates:
\begin{equation}
    S^* = OPTIMIZE(\hat{S}, \hat{K}, p_{s}(S)),
\end{equation}
where $\hat{S}$ - pooled site points, $\hat{K}$ - pooled discrete field, $p_{s}(S)$ - modelling data of original grid.

Our method uses $\Theta(S, K)$ as an initial approximation of $S^{*}$ and $K^{*}$ but optimizes $\hat{S}$ further, while averaged permeabilities $\Hat{K}$ are fixed.

\textbf{Optimization problem.}
$OPTIMIZE$ implements an optimization loop.
Let $e = RMSE(p_{s}(S), p_{s}(S^*))$ be the measure of closeness of the two time series.
Lower values of $e$ correspond to better fit at the measurement point.
If we manage to reduce the number of grid points and optimize their locations while keeping $e$ small, then we solve our main problem: we will be able to run simulations much faster without the loss of quality.
This makes our method self-adaptive because it doesn't need any data and controls itself.

For better readability, we assumed that there is just one measurement point, however, arbitrary number of points can be used in this approach.
Later we show application of our method in scenarios with multiple measurement points.

Finally, we formulate the following optimization problem:
\begin{equation}
\begin{aligned}
\operatorname*{min}_{S^*} MSE(p_{s}(S), p_{s}(S^*))
\end{aligned}
\end{equation}
We apply a stochastic minimization algorithm for this problem.
The number of optimization steps is a hyperparameter.
The computational graph includes differentiable Voronoi tessellation, FV solver and the loss function.
Backward pass allows to analytically compute partial derivatives of the loss function with respect to the site points.
Then the update step moves the site points so as to minimize the loss function.
As the output of the method we obtain the new position of the site points that represent the original physical system.

\section{Experiments and Analysis}

\subsection{Source specification}

In subsurface modelling, typical drivers for fluid flow
are injecting (sources) and producing wells. Both types of wells
can be described as a point source with its intensity proportional
to the difference between the pressure at the wellbore
and the pressure near the well. When such a source is discretized,
we receive the following expression,
\begin{equation}
    f_i^k =  
     \begin{cases}
      c_\alpha (p_{bh, \alpha} - p_\alpha^k) & \text{if} \, i=\alpha,\\
      0 & \text{otherwise,}
    \end{cases}      
    \label{eq:well_term}
\end{equation}
where $\alpha$ is the cell where the well is located, $c_\alpha$ is a parameter
depending on well size and permeability, and $p_{bh, \alpha}$ is the pressure 
at the wellbore.

In the presentation below, we slightly abuse the above definitions and use sink point/measurement point interchangebly.

\subsection{Varying the degree of reduction}

\begin{figure*}[hbt!]
    \center
    \begin{tabular}{cc}    
    \includegraphics[width=0.3\paperwidth]{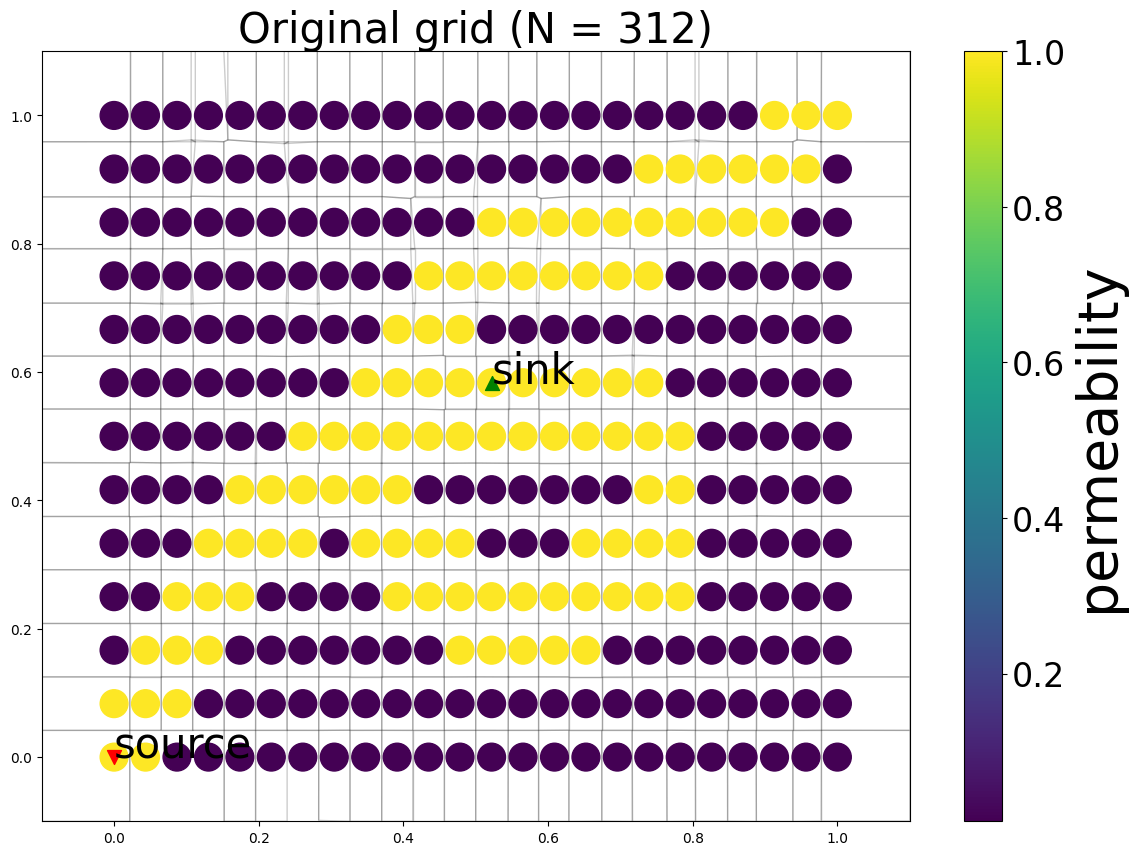} & 
    \includegraphics[width=0.3\paperwidth]{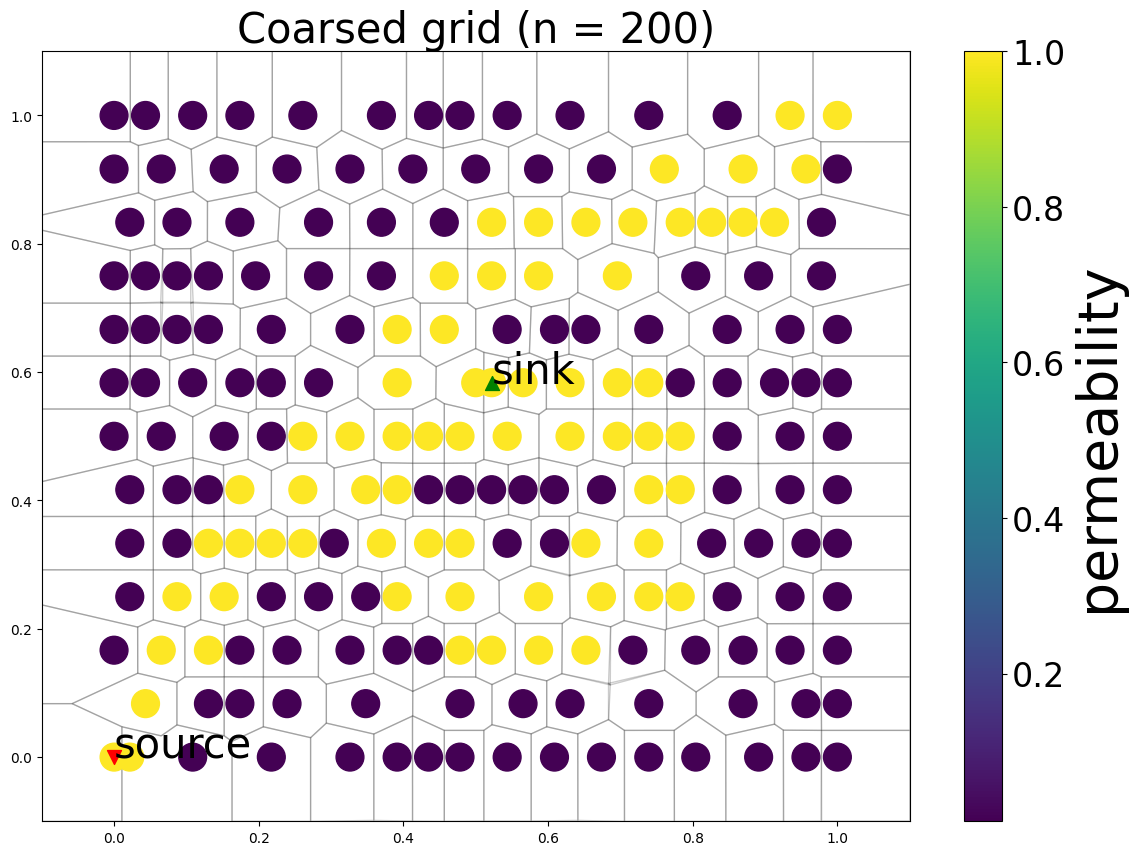} \\
    \includegraphics[width=0.3\paperwidth]{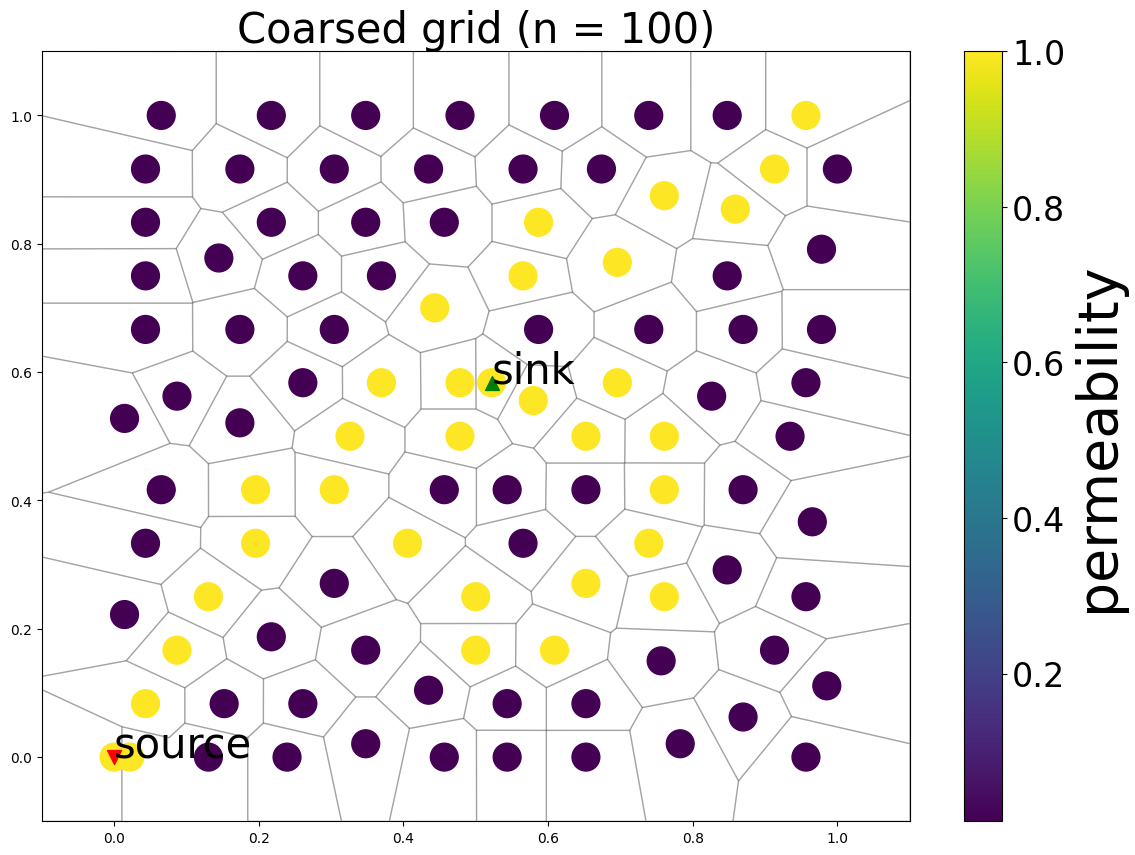} &
    \includegraphics[width=0.3\paperwidth]{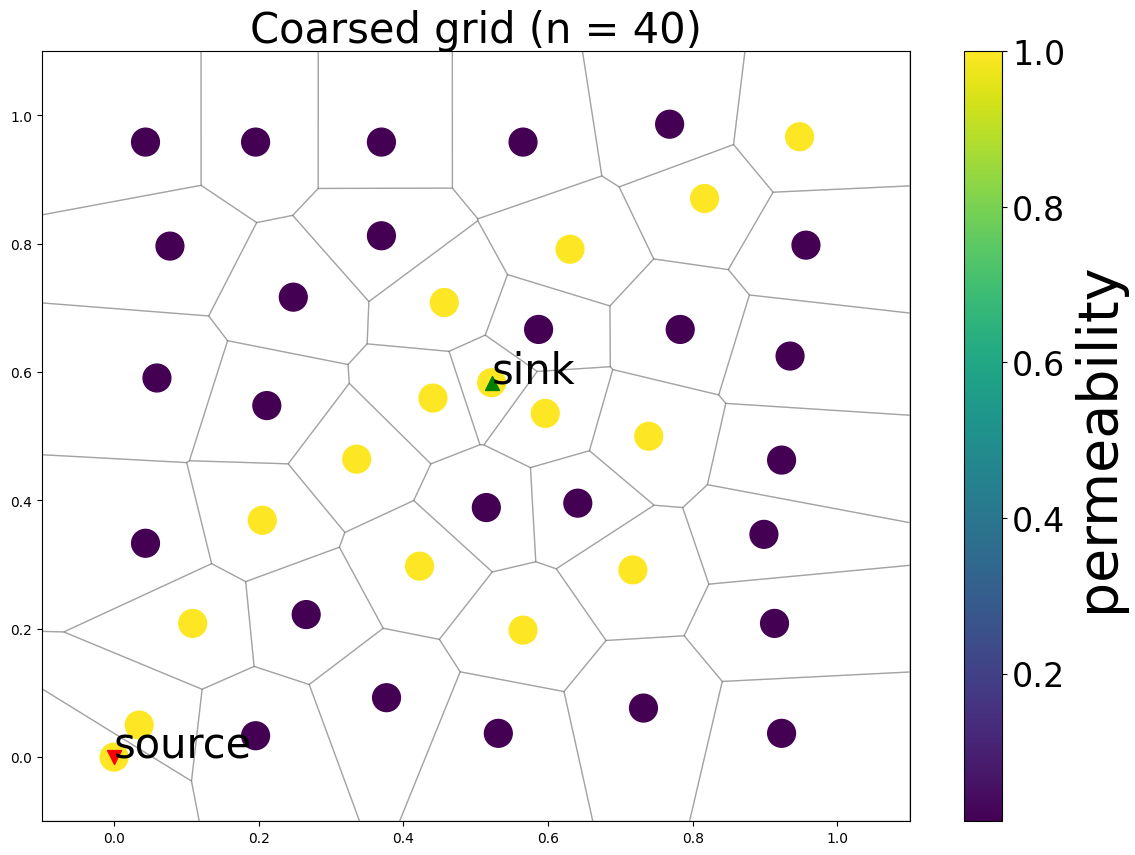}
    \end{tabular}
    \caption{Voronoi tessellation and discrete permeability field corresponding to different degrees of reduction. First we have an original point cloud, permeabilities, and $[-0.1, 1.1]^2$ boundary which together form the input data. We apply $k$-means clustering + mean pooling in clusters to the input data with different degrees of reduction: $\frac{200}{312}$, $\frac{100}{312}$, $\frac{40}{312}$. The figure shows the series of coarsened grids. The color represents the permeability. Source and sink points are kept untouched during coarsening.}
    \label{coarsening}
\end{figure*}

\begin{figure}[hbt!]
\begin{center}
\includegraphics[width=\columnwidth]{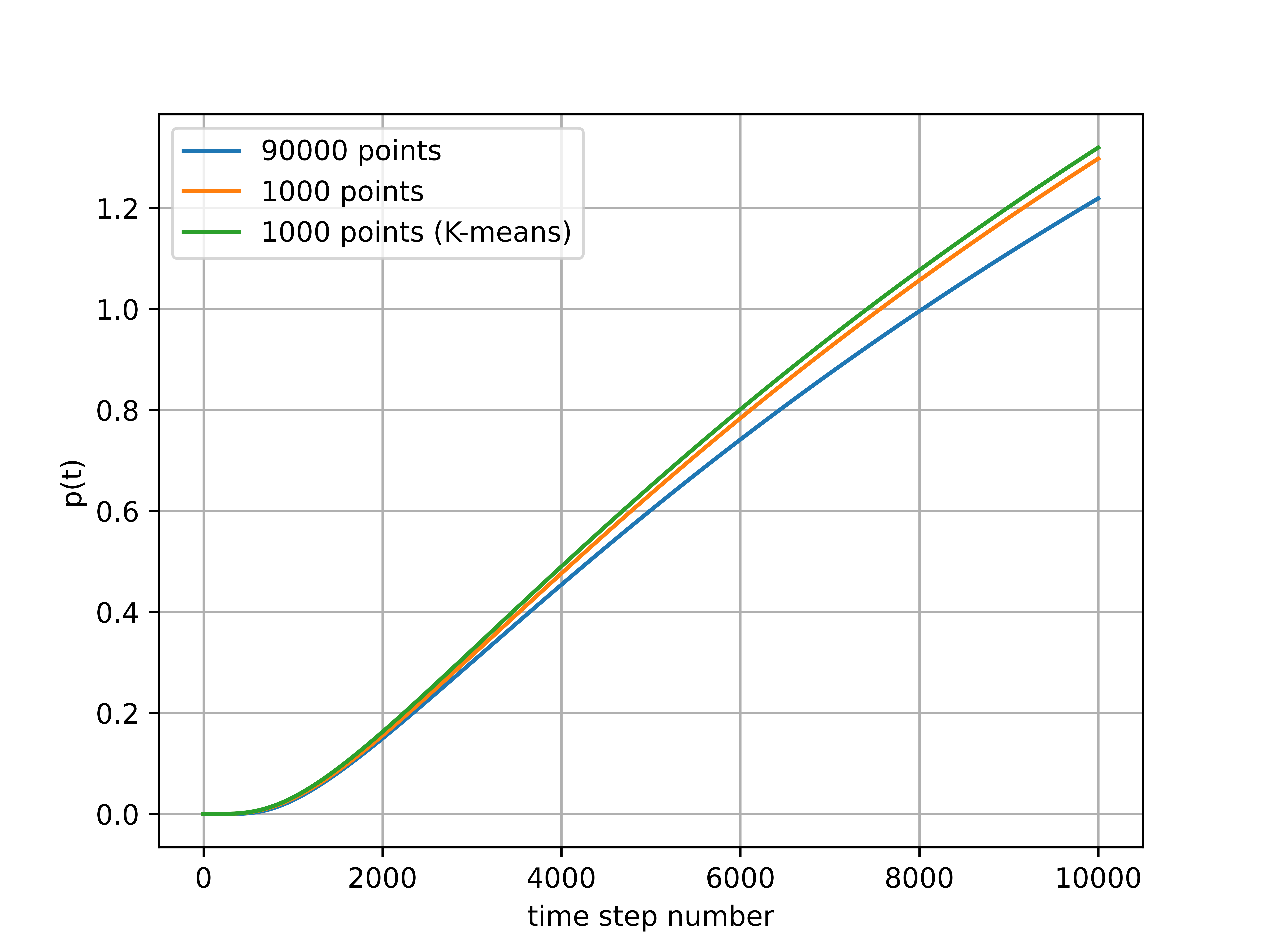}
\caption{The result of 90000$\to$1000 coarsening after 60 epochs. Our method proves ability to handle large point clouds in differentiable way, demonstrates good agreement with ground truth and improves $k$-means + average pooling result. After such coarsening simulation speed is 34x faster}
\label{great_coarsening}
\end{center}
\end{figure}

We define the degree of reduction as $r=n/N$.
As a first scenario we use the scenario we call \textbf{loop}.
It represents an evenly spaced point cloud $S$ ($N=312$) and a discrete permeability field $K$.
$K$ consists of \{0.1, 1\}.
In fact this scenario is much harder to coarsen comparing with a smooth permeability field because for some $\Theta$ it may disconnect the $K=1$ subset hindering the pressure propagation.
We choose different degrees of reduction to coarsen this grid (see Fig.~\ref{coarsening}).
Then we put all these grids into the FV simulator for $m=10^4$ steps and save the 
modelled $p(x, y, t)$ at the sink point.
As a result we have a pressure series vector $p_{s} \in \mathbb{R}^m$.

Fig.~\ref{coarsening} demonstrates the influence of $r$ on the $p_{s}$.
We set the following parameters for the modeling: $m=10^4$, $\tau=10^{-4}$, $p_{bh, src} = 1~Pa$, $c_{src}=0.5 m^3 /s $.
Than we compare $p_{sink}$ for $0<t<T$ for the four scenarios: $r$ = \{1.0, 0.75, 0.5, 0.25\}.
For all the simulations we use differentiable finite volume Solver which solves \eqref{eq:parab} for $p(x, y, t)$.
We set a $[0, 1]^2$ boundary.

\begin{figure}[hbt!]
\begin{center}
\includegraphics[width=0.9\columnwidth]{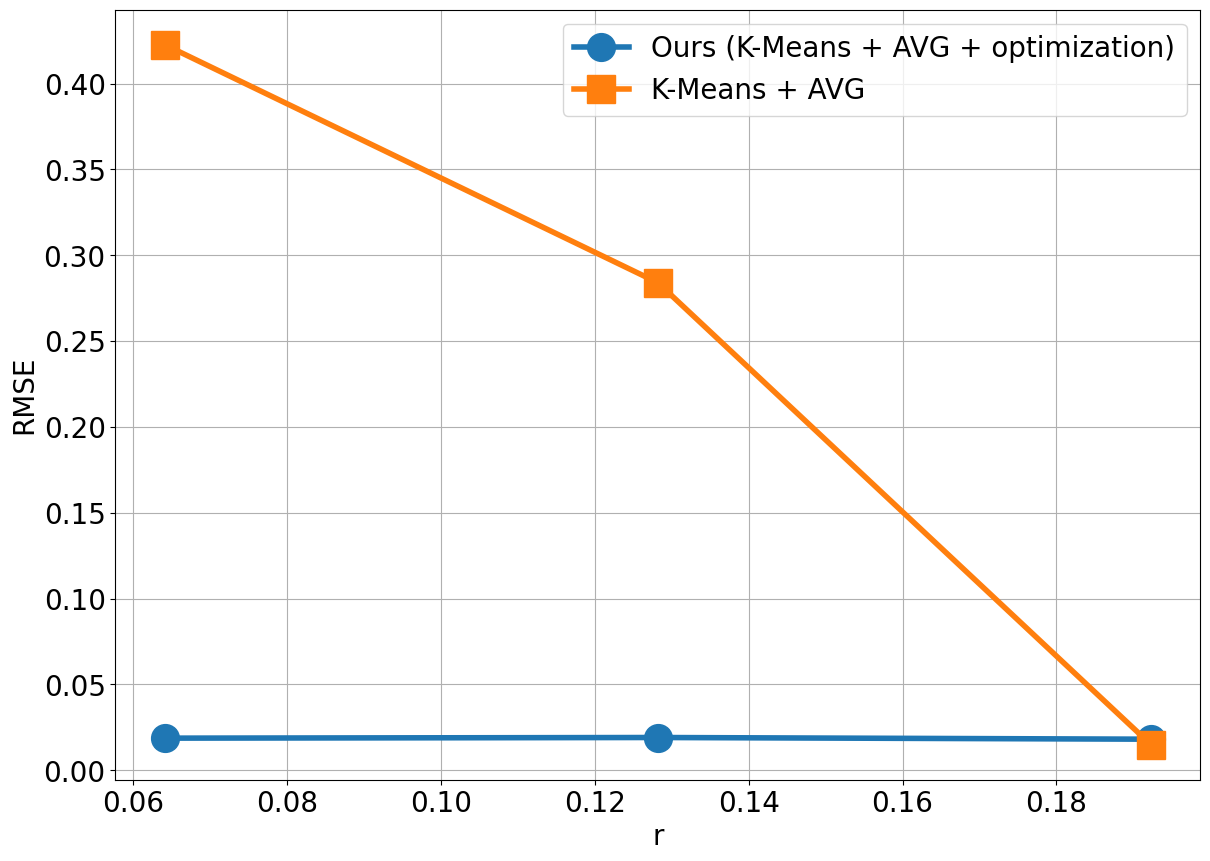}
\caption{The result show our method allows to make coarsening of much lower degree but preserving the modeling quality.}
\label{comparison_rmse}
\end{center}
\end{figure}

\begin{figure*}[hbt!]
    \begin{tabular}{cc}    
    \includegraphics[width=\columnwidth]{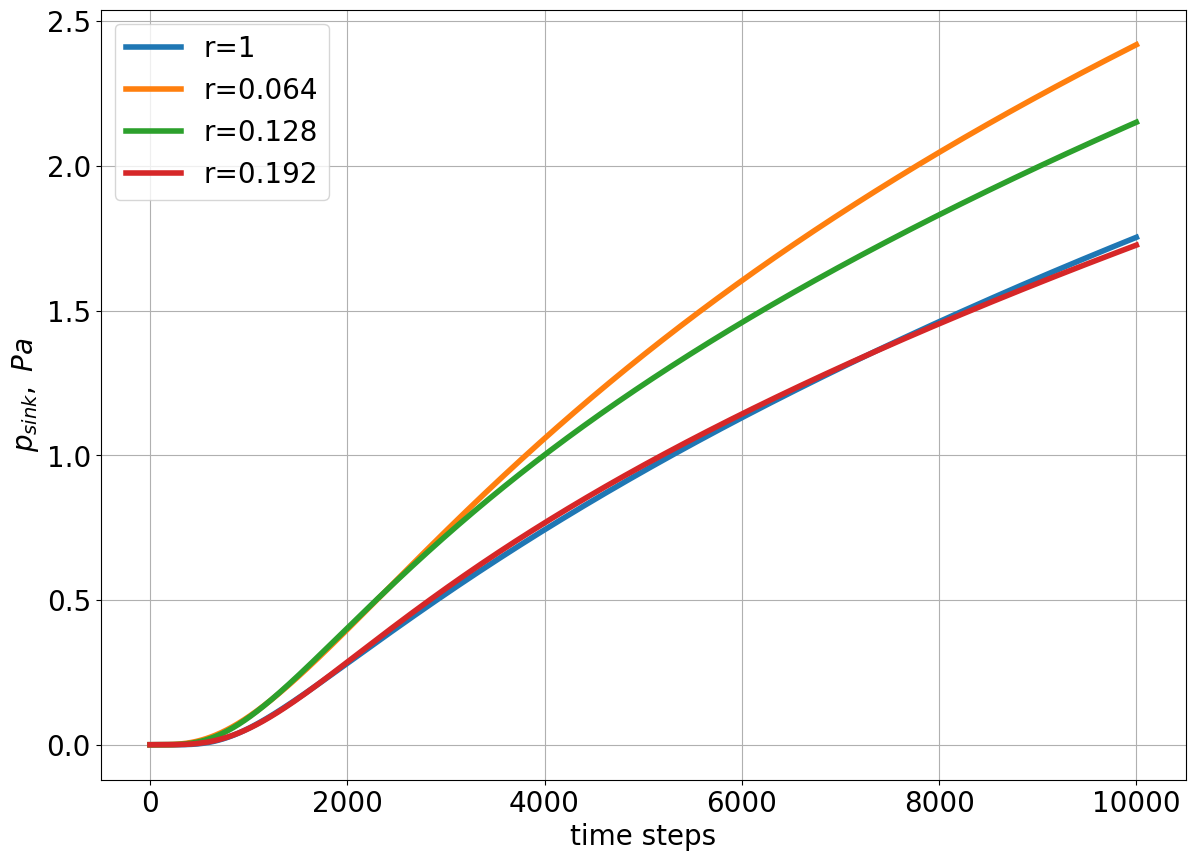} & 
    \includegraphics[width=\columnwidth]{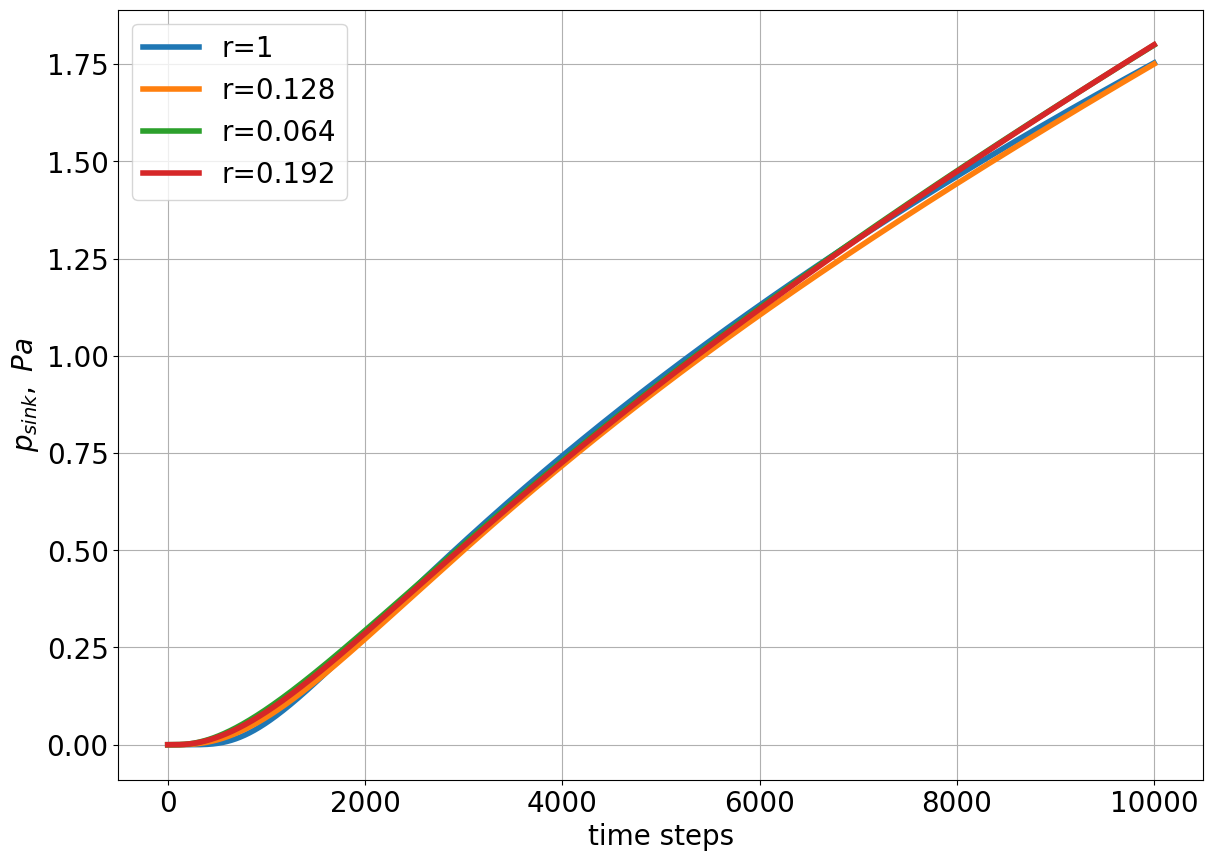} 
    \end{tabular}
    \caption{a) Comparison of $p_{s}$ for different degrees of reduction for \textit{loop} scenario. For coarsening \textbf{$k$-means + averaging} are used. b) Comparison of $p_{s}$ for different degrees of reduction for \textit{loop} scenario by using \textbf{our method}. Optimization is done for 20 epochs. Adam optimizer. Learning rate $10^{-3}$.}
    \label{ours_vs_kmeans}
\end{figure*}

Keeping the described setting we compare the $\Theta$ which use $k$-means for clustering and average pooling vs. our method.
The results of the experiment are shown on the Fig.~\ref{ours_vs_kmeans}.
The results show that our method allows to significantly improve the modeling, compared to the original pressure series.
All the series are much close to the ground truth one.
The comparison of RMSE for this experiment is presented on the Fig.~\ref{comparison_rmse}
For this experiment we've done optimization for 20 epochs and modelled $10^4$ time steps.

\begin{figure*}[hbt!]
    \begin{tabular}{cc}    
    \includegraphics[width=\columnwidth]{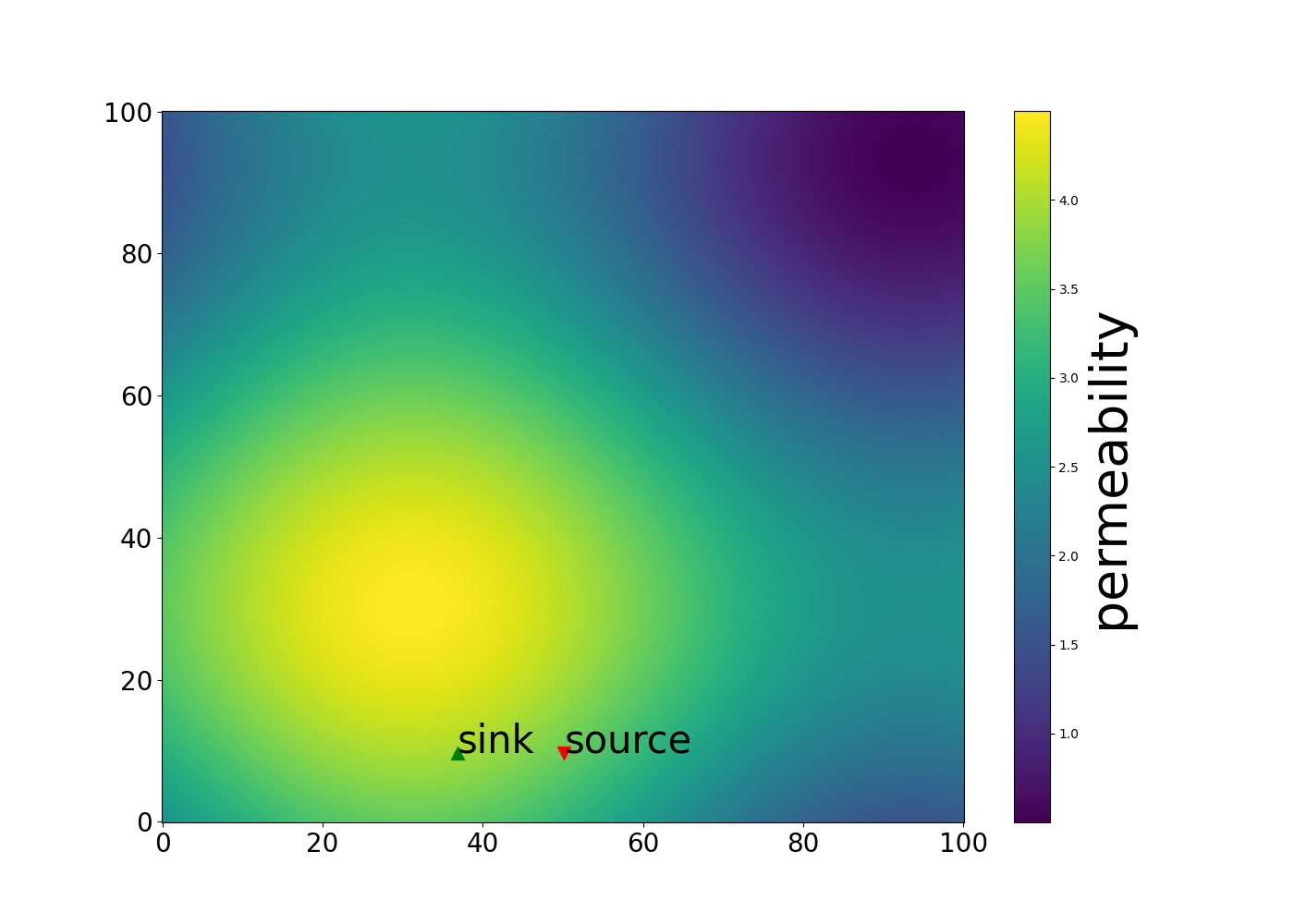} & 
    \includegraphics[width=\columnwidth]{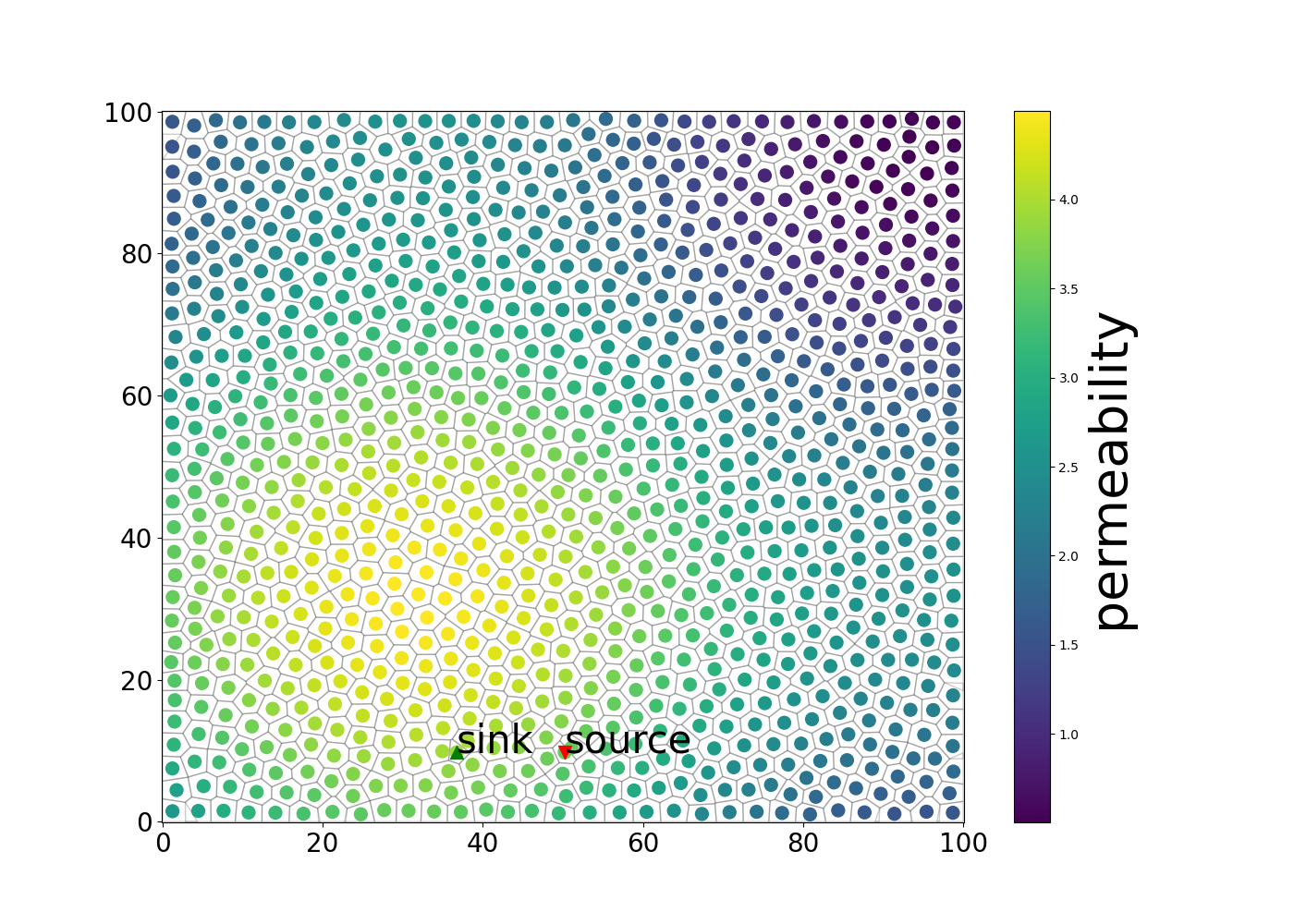} 
    \end{tabular}
    \caption{Voronoi tessellation corresponding to 90000$\to$1000 coarsening.  Left plot represents original point cloud, right - coarsened point cloud by our pipeline. The color represents the permeability. Source and sink points are kept untouched during coarsening.}
    \label{largescale_coarsening}
\end{figure*}

\subsection{Hyperbolic equation case}

The wave equation, an example of a hyperbolic PDE, models phenomena such as sound or other types of waves propagating through various media.
In fluid dynamics, the wave equation can be used to describe pressure waves in compressible fluids. 
When disturbances occur in the fluid, they generate pressure waves that propagate according to the wave equation:
\begin{equation}
    \frac{\partial^2 p}{\partial t^2} - c^2 \nabla^2 p = f
\end{equation}
where $c$ is speed of sound.

We have adapted our solver for the wave equation. 
The modified solver uses the same finite volume discretization scheme as for the parabolic equation, but we changed the update formula:
\begin{equation}
p^{k+1} = -p^{k-1} + 2p^k + \tau^2 D^{-1} A p^k + \tau^2 f^k,
\end{equation}
%In this experiment, the right-hand side term is defined as:
%\begin{equation}
%f^k = \left( \frac{2 \pi^2}{L^2} - 1 %\right) \cos\left( \frac{\pi x}{L} %\right) \cos\left( \frac{\pi y}{L} %\right) \sin(\tau k)
%\end{equation}
%
%where:
%\( L = 10 \) is the length scale.

Our pipeline has shown the ability to handle the wave equation well.
The results of experiments for wave equation are described in Appendix~\ref{app:wave} \footnote{https://youtu.be/4rI0vf2zYCQ}.

\subsection{Multiple measurement points and predicting ability}

One of the most important application of model order reduction is oil reservoir modeling because usually computational grids for these models have millions of cells.
Reservoir modeling is a very costly computation and reduction of the computation domain may save a lot of effort.
A lightweight model of an oil reservoir is called a proxy model.
In fact coarsened coordinates and permeabilities are the proxy model of some physical field.

To test the ability of the method to get coarsened grids good enough for simulation beyond the time used for coarsening, we implement the traditional machine learning approach by splitting the train and test time periods.

After coarsening we run simulation again but for a larger number of steps and check the predicting ability of the proxy model.
Experiments demonstrate that coarsened grid is able to produce accurate simulation compared to the original grid. The details of the experiment are described in Appedix~\ref{app:real_world}.
The video of the process of optimization is available online\footnote{https://youtu.be/JBC1v8xSE58}.

\subsection{Scalability}

\begin{table*}[t]
\caption{Performance metrics of the graph simulator for different sizes of the task.}
\label{tab:performance-table}
\vskip 0.15in
\begin{center}
\begin{small}
\begin{sc}
\begin{tabular}{lcccr}
\toprule
Number of Points & Time (minutes) & Peak RAM (GB) \\
\midrule
90,000                    & 17             & 2.8                  \\
250,000                   & 51              & 3.2                  \\
1,000,000                 & 239              & 4.9                  \\ 
\bottomrule
\end{tabular}
\end{sc}
\end{small}
\end{center}
\vskip -0.1in
\end{table*}

\begin{table*}[t]
\caption{Performance metrics of one epoch of the optimization loop with different scenarios (10000 timesteps).}
\label{tab:performance_metrics}
\vskip 0.15in
\begin{center}
\begin{small}
\begin{sc}
\begin{tabular}{lcccr}
\toprule
Scenario & Time (min) & Peak RAM (GB) \\
\midrule
90000$\to$1000 & 0.9 & 7.1 \\
90000$\to$500 & 0.7 & 5.7 \\
90000$\to$100 & 0.45 & 3.1 \\
\bottomrule
\end{tabular}
\end{sc}
\end{small}
\end{center}
\vskip -0.1in
\end{table*}

An essential aspect of assessing the efficacy of the graph-based simulator and general pipeline is its scalability, particularly as the size of the point cloud increases.
This section presents a performance analysis, focusing on the execution time and memory consumption, across varying sizes of point clouds. 
The experiments were conducted on Google Colab, utilizing two cores of an Intel(R) Xeon(R) CPU @ 2.20GHz and 12.7 GB of RAM, to ensure a consistent and replicable environment.

First, we perform the benchmarking of the graph-based simulator. 
Point clouds were tested on several scales and these scales were chosen to represent a broad range of potential real-world applications, from moderate to highly dense point clouds. 
Execution time and RAM consumption were measured for each scale, providing information on simulator efficiency and resource utilization.  
For all experiments presented here, we used synthetic permeability generated by the following function: $sin(ax) + sin(by) + 2.5$ where $a = b = 0.05$. 
The coordinates $x$ and $y$ were systematically varied in a continuous domain from 0 to 100, with both axes subdivided into 300 equidistant intervals. 
This form of function leads to permeability in the range $[0.5, 4.5]$. 
Fig.~\ref{largescale_coarsening} shows the generated permeability field.  
The locations of the source and measurement points are arbitrarily selected, the source point is positioned at coordinates approximately [50.17,9.70], and the sink point is located at [36.79, 9.70]. 
Parameters $c_{src}=1~m^3 /s $, $p_{src}=50$ $Pa , \tau =  0.005\ s. $ has been chosen for these experiments. 
We set a $[0, 100]^2$ boundary.

The results in the table~\ref{tab:performance-table} provide critical insights into the scalability of the simulator. 
Execution time and RAM consumption metrics are key indicators of the simulator's ability to handle large-scale data efficiently.  

Additionally, we measured the performance of the general pipeline. 
For this task, we tested several types of coarsening (number of points before$\to$after): 90000$\to$1000, 90000$\to$500, 90000$\to$100 and measured duration of one epoch of optimization described in section $General\ pipeline$. 

We note that overall execution time of the optimization step is dominated by simulator run on optimized (reduced) point cloud. 

Finally, we demonstrate and highlight the potential gains from using our approach. 
For illustrative purposes, we applied coarsening to the point cloud, reducing its size from 90,000 to 1,000 points using the $k$-means algorithm, mirroring the methodology detailed in the preceding subsection (see Fig. \ref{largescale_coarsening} ). 
This reduction, by a factor of 90, led to a dramatic decrease in the simulation execution time, by a factor of 34, resulting in an average computation time of approximately 0.5 minutes. 
Remarkably, after just 60 epochs, our model achieved an RMSE of approximately 0.042, signifying a notable improvement over the baseline $k$-means + average pooling result of 0.057. 
These results are visually depicted in Fig. \ref{great_coarsening}, highlighting the effectiveness of our approach.

\section{Conclusion, Limitations, and Future Work}
% 
% 3D
% heterogenos anisotrpic
% implicit
% history match
% Voronoi

We presented a new method to coarsen an unstructured computational grid
that provides much faster modeling without significant loss in accuracy.
The method is based on minimization of the misfit between time series modelled on coarse and fine grids.
Our method includes $k$-means clustering, autodifferentiation
and stochastic minimization algorithms.
The method makes use of the combination of the differentiable Voronoi tessellation with the differentiable finite volume solver that allows to pass gradient 
from the differentiable loss function to the input locations of the points.
In our experiments we reduced the problem size 10 times whilst preserving preserved the quality of modeling.

We tested our method using the important practical example of slightly compressible fluid flow in porous media. Our approach showed great flexibility and provided a 34x speedup while keeping the accuracy of fine grid modeling.

Our method is focused on a specific method for solving PDE but not specific types of PDEs.  
Our approach is applicable to any finite volume, two-dimensional explicit solver compatible with unstructured meshes—provided the differential operators can be represented through Voronoi tessellation. 
This is due to the fact that any pertinent numerical scheme can be represented as a computational graph over a cloud of points (the cell centers), which can then be spatially coarsened. 
The resulting coarser point cloud can again be used to build unstructured grid and calculate the differential operators, thereby maintaining the integrity of the solver's operations. 
Therefore, if a specific method of solver (finite volumes method in our case) can solve some kind of PDE, then our method can be also applied to this PDE type.

In our current work, the usage of the explicit Euler scheme was sufficient for demonstration of the general concept of coarsening and application of differentiable Voronoi tessellation. 
We are keen on investigating implicit schemes in the subsequent studies.

%The method allows to get sparsified representation of the computational unstructured grid and allows to perform faster computations.
%We present the end-to-end differentiable solution that allows to pass gradient from the differentiable loss function to the input locations of %the points.
%Then we move points until convergence.
%We compared our method with alternative methods for the sparsification of the point cloud.
%We presented differentiable Finite Volume Solver and differentiable version of the Voronoi tessellation.

\section*{Impact statement}
\textbf{Community impact.}
The work opens up the possibility to use differentiable simulators together with adaptive coarsening techniques in order to reduce the size of synthetic and real-world problems.

\textbf{Industry impact.}
The filtration equation discussed in this paper is used to model groundwater as well as oil reservoir dynamics.
Modeling the movement of water underground allows us to better assess available reserves and build more environmentally friendly infrastructure. 
Our method allows us to speed up the calculation of the filtration equation, which allows us to solve problems related to the installation of wells and groundwater analysis.

Also our method contributes to petroleum engineering.
It allows to solve both forward and inverse problems using proxy models, obtained with our method.

\section*{Acknowledgements}

The work was supported by the Analytical center under the RF Government (subsidy agreement 000000D730321P5Q0002, Grant No. 70-2021-00145 02.11.2021).

We also acknowledge insightful discussions with Dr. Alexander Korotin and Petr Mokrov (Skoltech).

\bibliography{bibliography}
\bibliographystyle{icml2024}

%%%%%%%%%%%%%%%%%%%%%%%%%%%%%%%%%%%%%%%%%%%%%%%%%%%%%%%%%%%%%%%%%%%%%%%%%%%%%%%
%%%%%%%%%%%%%%%%%%%%%%%%%%%%%%%%%%%%%%%%%%%%%%%%%%%%%%%%%%%%%%%%%%%%%%%%%%%%%%%
% APPENDIX
%%%%%%%%%%%%%%%%%%%%%%%%%%%%%%%%%%%%%%%%%%%%%%%%%%%%%%%%%%%%%%%%%%%%%%%%%%%%%%%
%%%%%%%%%%%%%%%%%%%%%%%%%%%%%%%%%%%%%%%%%%%%%%%%%%%%%%%%%%%%%%%%%%%%%%%%%%%%%%%

\newpage
\appendix
\onecolumn

\section{Example of work of differentiable finite volume solver (Darcy equation)}

\begin{figure*}[hbt]

    \includegraphics[width=0.35\textwidth] {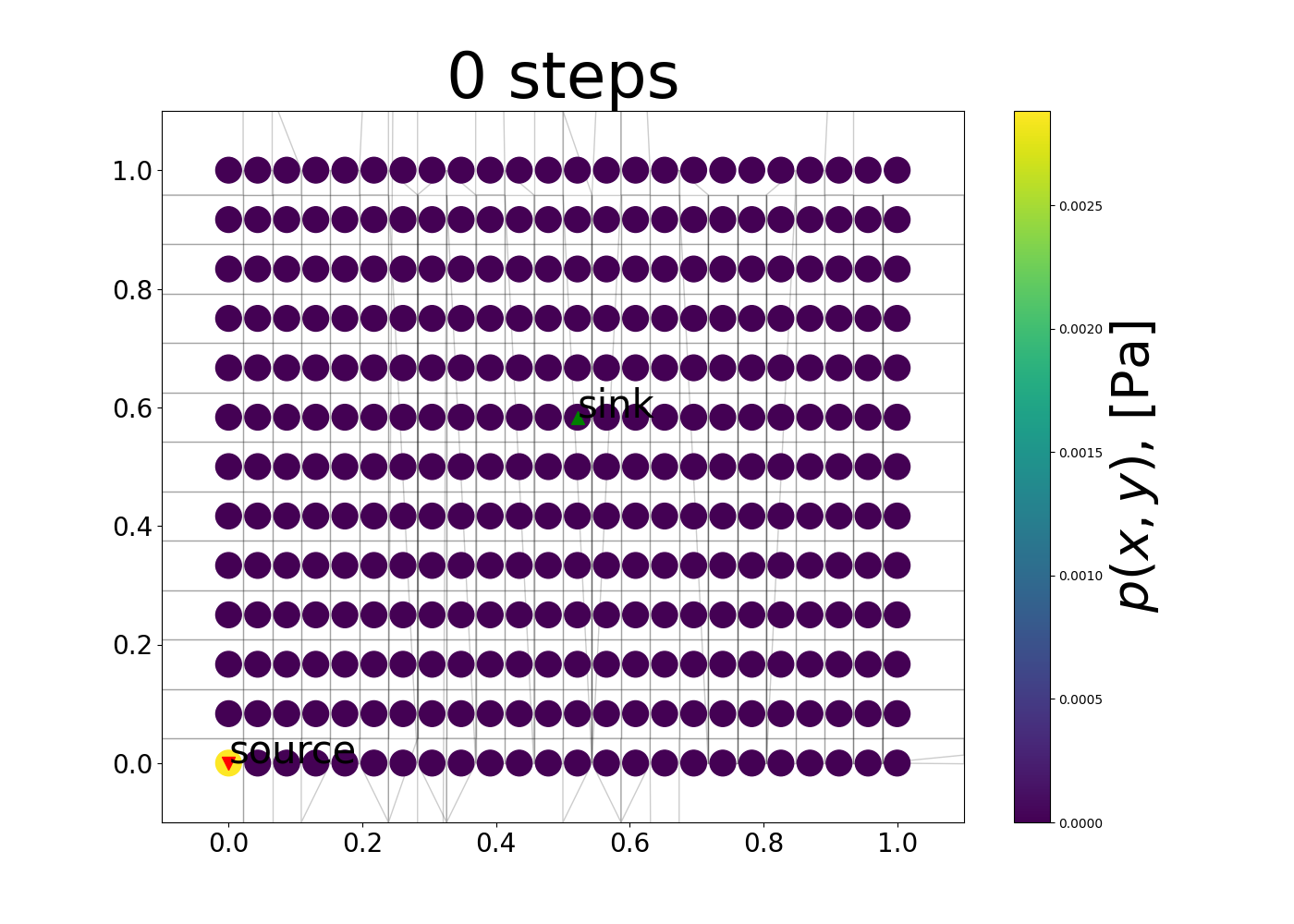} % 
    \includegraphics[width=0.35\textwidth]{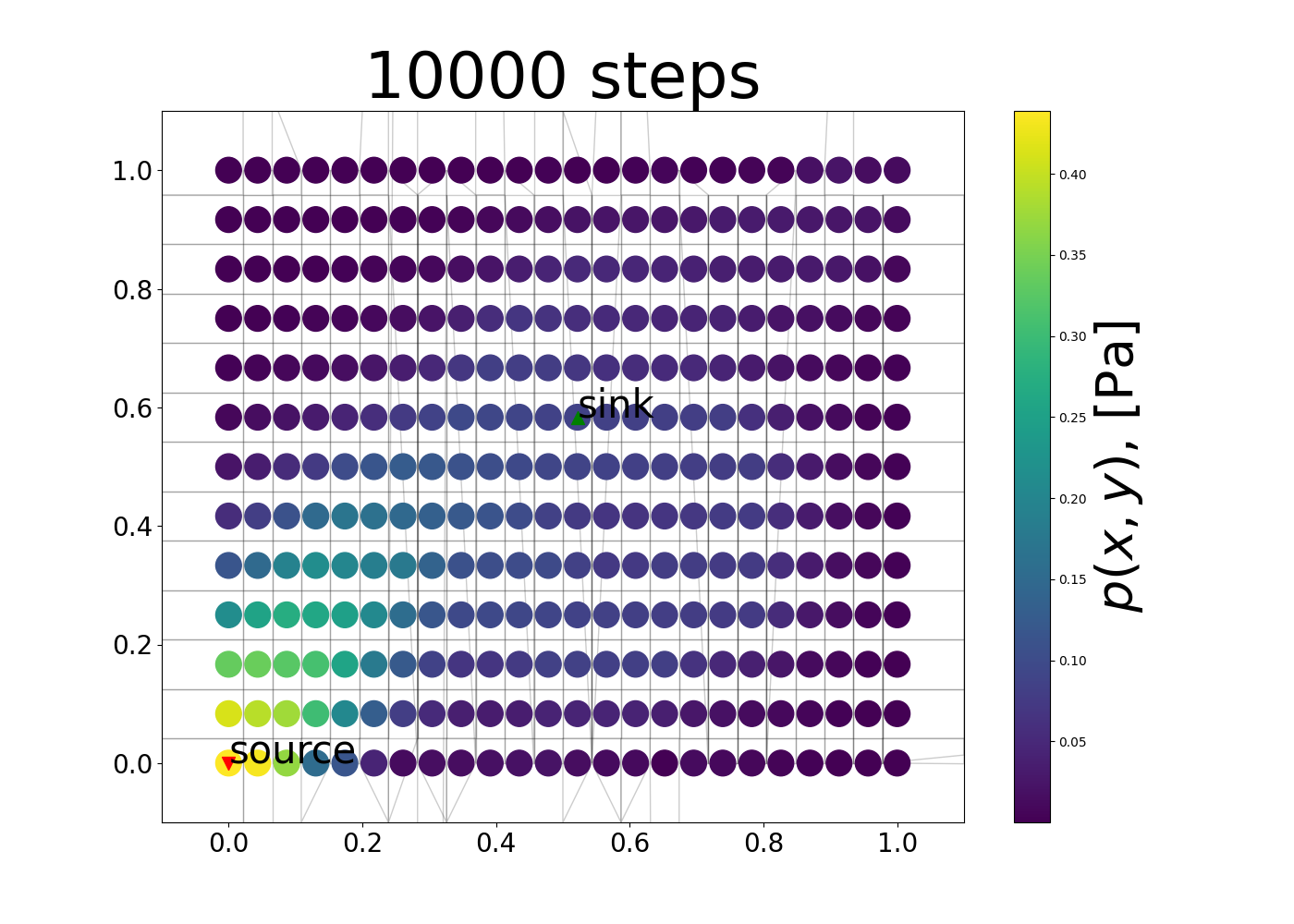} % 
    \includegraphics[width=0.35\textwidth]{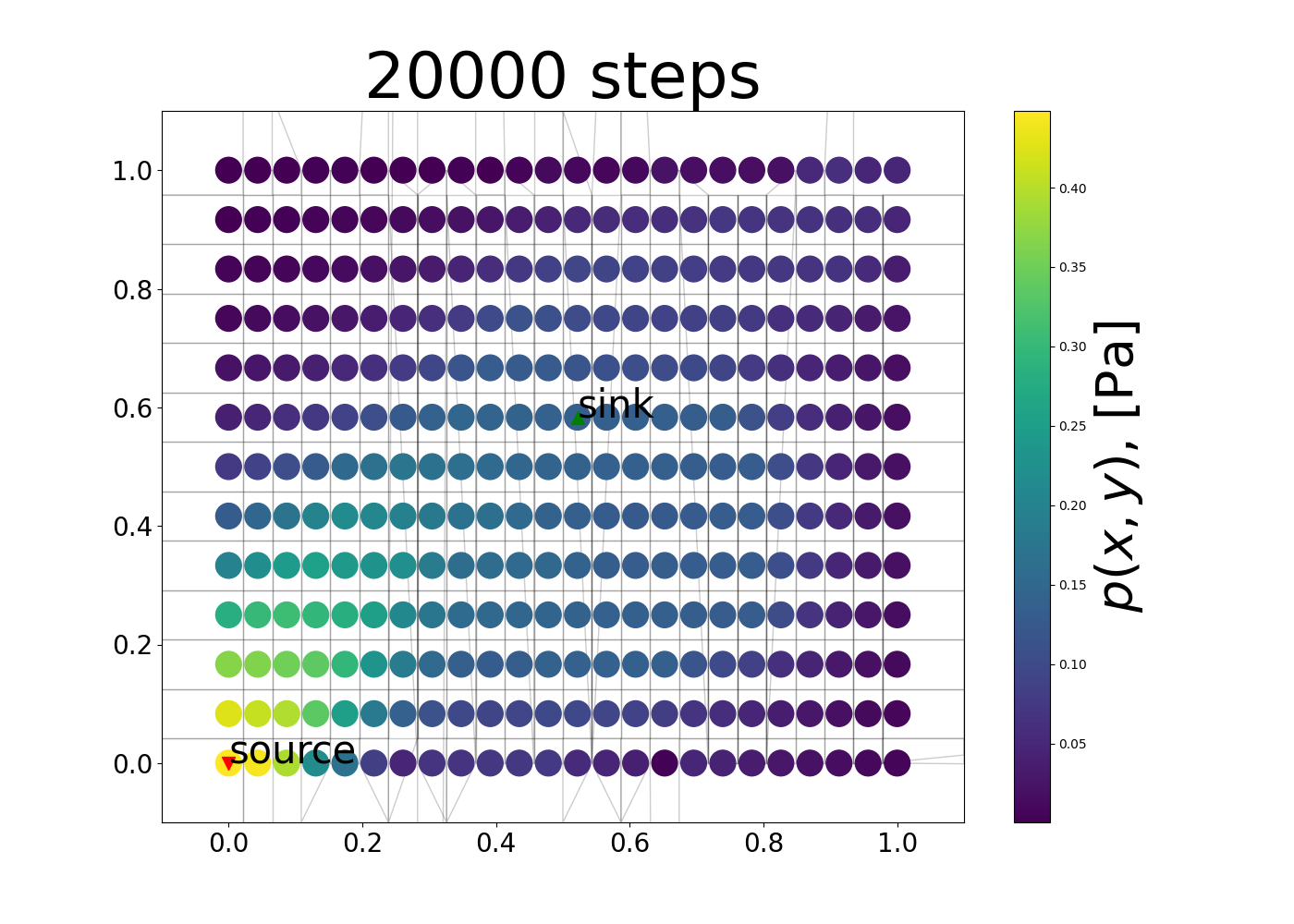} %
    \includegraphics[width=0.35\textwidth]{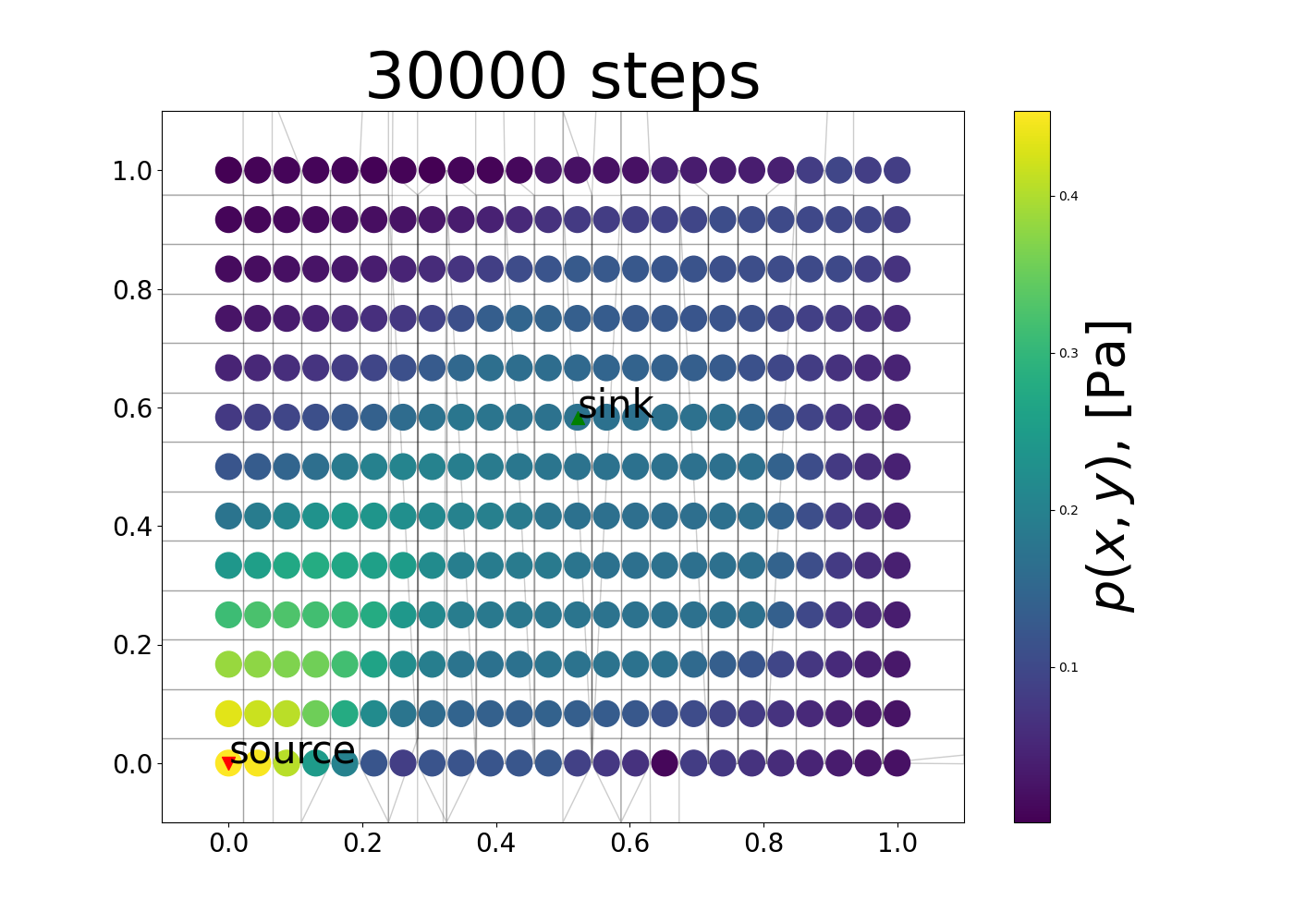} %
    \includegraphics[width=0.35\textwidth]{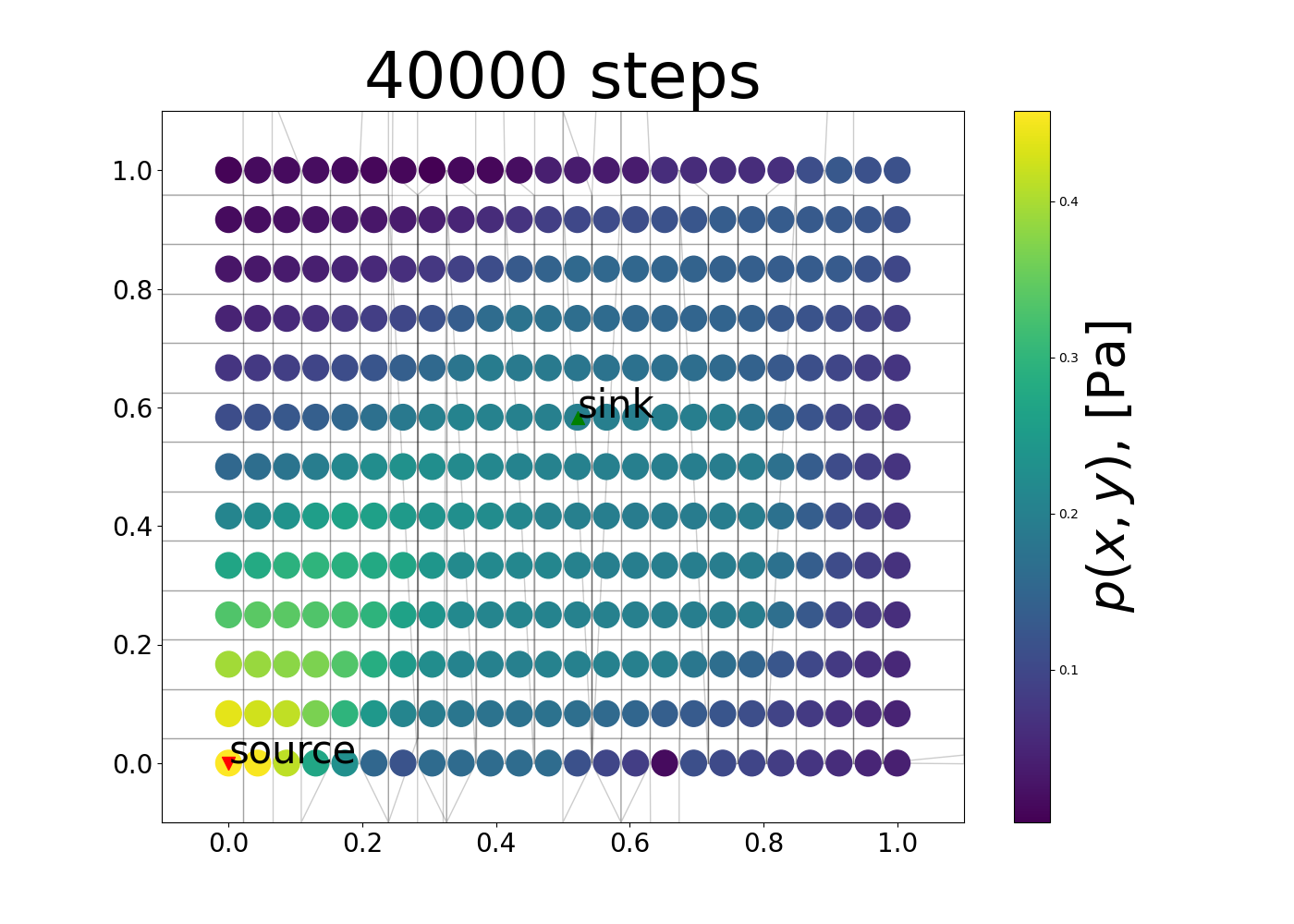} %
    \includegraphics[width=0.35\textwidth]{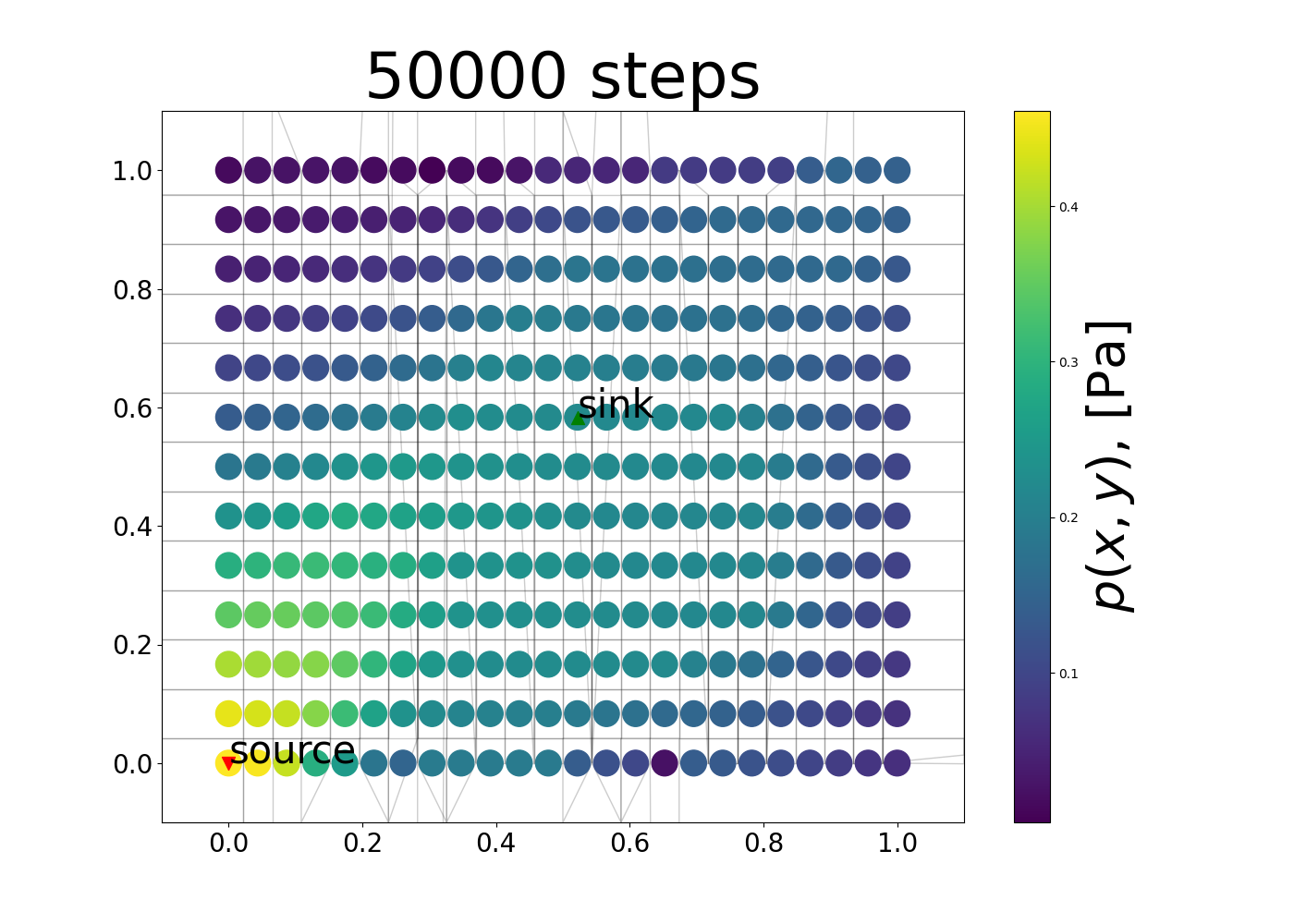} %
    \includegraphics[width=0.35\textwidth]{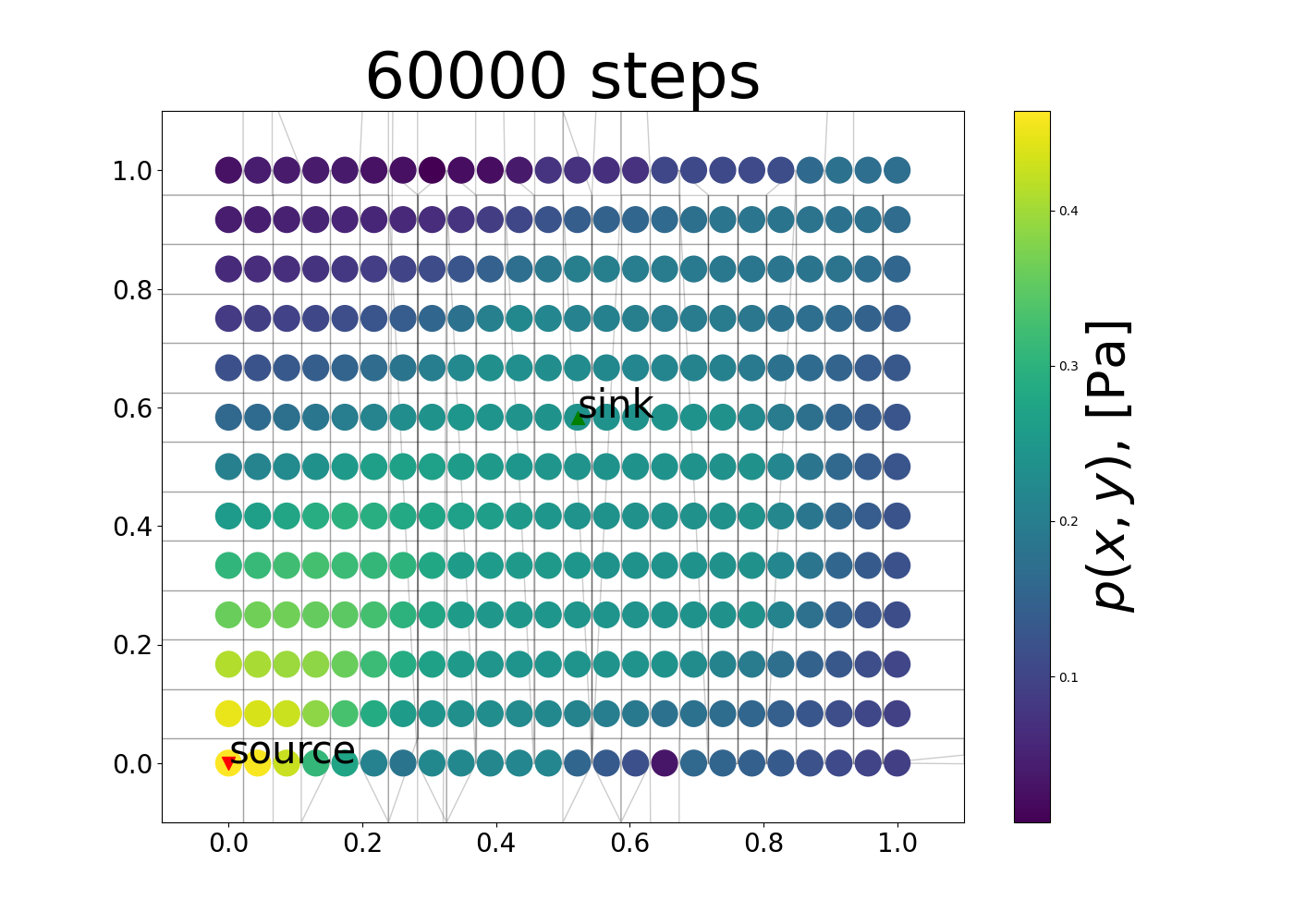} %
    \includegraphics[width=0.35\textwidth]{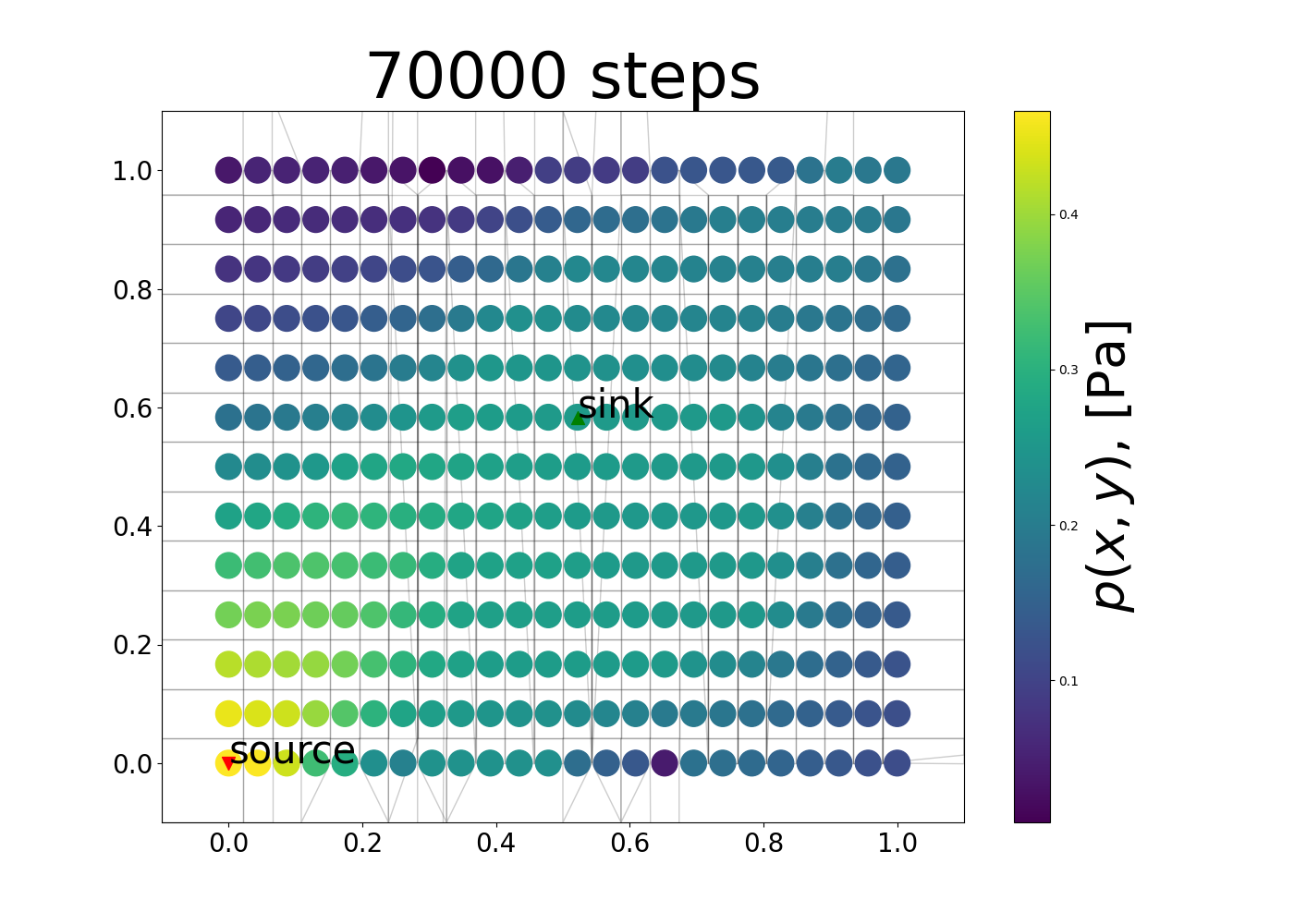} %
    \includegraphics[width=0.35\textwidth]{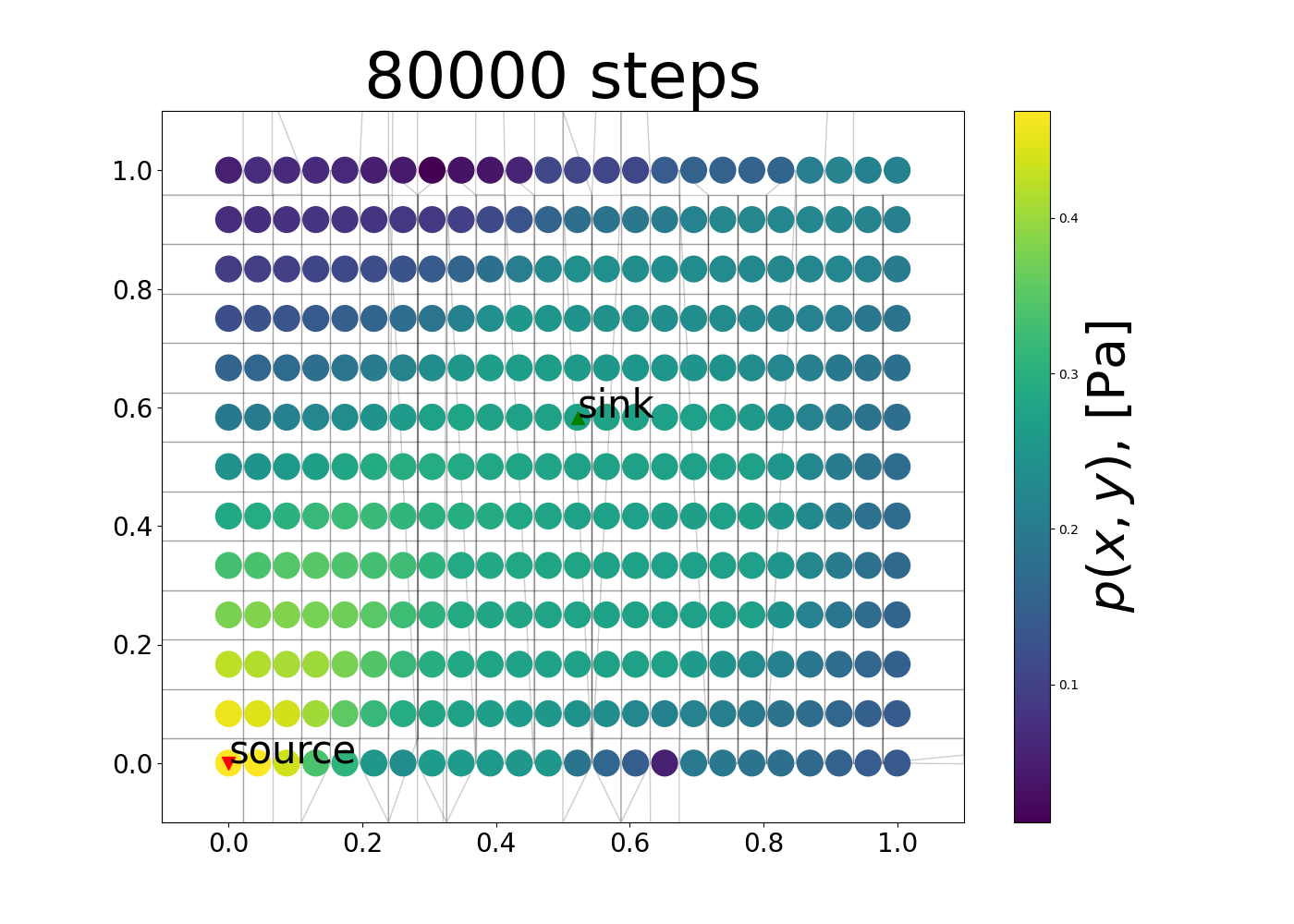}

    \caption{The work of the Differentiable finite volume solver for Darcy equation. The values of $p(x,y)$ is shown in every node of the graph. Graph is the Voronoi graph. An edge $i, j$ exists in this graph if $V_i$ and $V_j$ are adjacent. The source point injects the fluid into the subsurface domain. As the parameters for the run we used $c_{src}=1~m^3 /s$, $p_{src}=0.5$ $Pa$, $\tau=0.0001$. Points are evenly spaced because it's the original point cloud $S$ and it's not yet coarsened.}
    \label{pressure}
\end{figure*}

\section{Results for the wave equation}
\label{app:wave}

Here, we present our experiments with self-supervised coarsening
applied to the acoustic wave equation.
We considered a domain \([0, L] \times [0, L]\), where \( L = 10 \).
The grid was generated by distributing evenly 400 points. 
The speed of sound, $c(x,y)$, was equal to 1. There was no 
energy source, $f = 0$.

The initial conditions for the pressure field was a Gaussian-like distribution, while its temporal derivative was zero,
\begin{equation}
p(x, y, 0) = \exp\left(-\frac{(x - 5)^2}{2} - \frac{(y - 5)^2}{2}\right), \quad
p_t(x, y, 0) = 0.
\end{equation}
%
%The initial conditions are given by:
%
%\begin{equation}
%p \bigg|_{t=0} = 0
%\end{equation}
%
%\begin{equation}
%\frac{\partial p}{\partial t} %\bigg|_{t=0} = \cos\left( \frac{\pi x}{L} %\right) \cos\left( \frac{\pi y}{L} \right)
%\end{equation}
\begin{figure*}[hbt]
    \includegraphics[width=\textwidth]{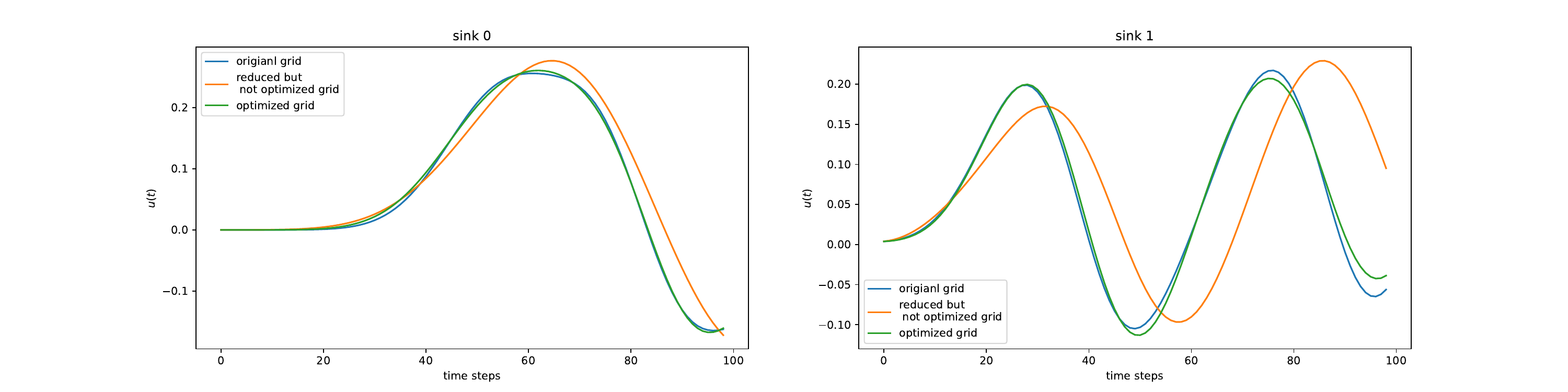}
    \caption{Results of the our approach applied to wave equation. Our method is able to adapt the grid to minimize the divergence of the sinusoidal signals in two sink points}
    \label{fig:wavesolver-results}
\end{figure*}
The video demonstrating work of our solver in this case is available online\footnote{https://youtu.be/SCBj1Jcvd7Y}.

Our methodology for this experiment mirrors the approach taken with previous cases. We applied our approach to coarsen the computational grid from 400 to 60 points. The time step, $\tau$ was equal $0.001\ s$ in these experiments. The domain was $[0, 10]^2$. The results are demonstrated in the Fig. \ref{fig:wavesolver-results}. The video demonstrating the experiment is available online\footnote{https://youtu.be/4rI0vf2zYCQ}.

The successful application of our coarsening algorithm to the wave equation broadens the spectrum of problems where our approach can be confidently applied.

\section{Stability Analysis}
\label{sec:stability}

Stability analysis of the explicit scheme \eqref{eq:fvm} 
shows that the time step should be limited by the fraction,
\begin{equation}
    \tau \leq \frac{2}{\lambda_{max}},
\end{equation}
where the denominator is largest eigenvalue of matrix $D^{-1}A$. The later could be estimated 
with the Gershgorin disk theorem,
\begin{equation}
\lambda_{max} \leq 2 \, C \,  \frac{K_{max}}{V_{min}},
\end{equation}
where  $V_{min}$ is the smallest cell area,  $K_{max}$  is the largest permeability, 
and $C$ is a dimensionless constant, which depends on 
grid geometry and topology.
Consequently, we arrive to a sufficient condition of stability, 
\begin{equation}
    \tau \leq \frac{V_{min}}{C \, K_{max}}. 
\end{equation}

\color{black}

\section{Petroleum engineering example}
\label{app:real_world}

\begin{figure}[hbt]
    \centering
    \includegraphics[width=\textwidth]{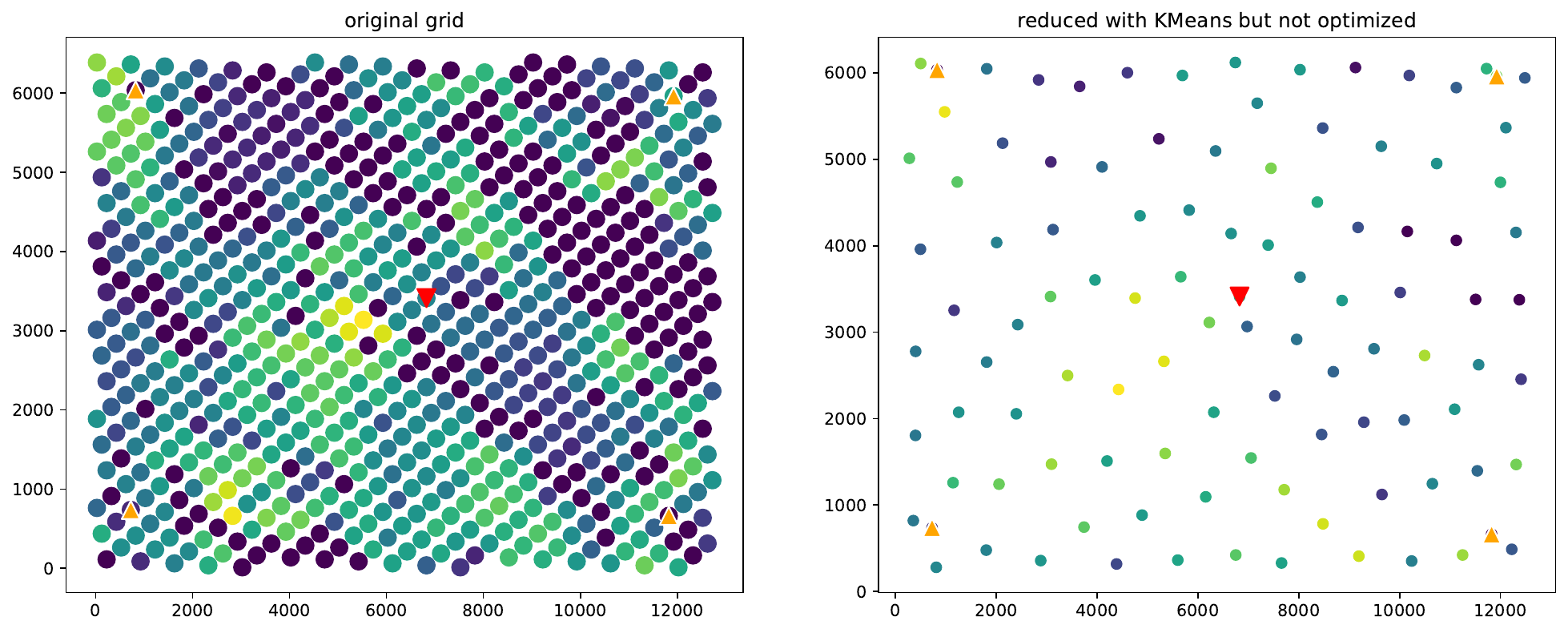}
    \caption{Real-world data experiment with 4 measurement point. We conduct an experiment with 4 measurement points and real-world data permeability field. We simplified the permeability field beforehand for visualization purposes.}
    \label{fig:4sinks_experiment setup}
\end{figure}

The governing equation for modelling an oil reservoir is Darcy PDE \ref{eq:parab}. 
In this scenario injecting wells are sources and production wells are measurement points.
We need to control the oil debit for production wells.
Let's consider the permeability field of a real oil reservoir model (see~Fig.\ref{fig:4sinks_experiment setup}).
Permeabilities in this field ranged from $10^{-2}$ to $10^{2}$ mD.
Red triangle represents a source point.
Orange triangles represent measurement points. We set $p_{bh, src} = 10^3$ Pa.
We test the ability of our method to output a proxy model that is capable to predict the future oil debit.
First, we use our method to coarsen an unstructured grid.
We use $m$ = 10000.
After optimization we use the resulting grid and pass it again into the FV solver but for now we model it for $m=20000$ time steps.

\begin{figure}[hbt]
    \centering
    \includegraphics[width=\textwidth]{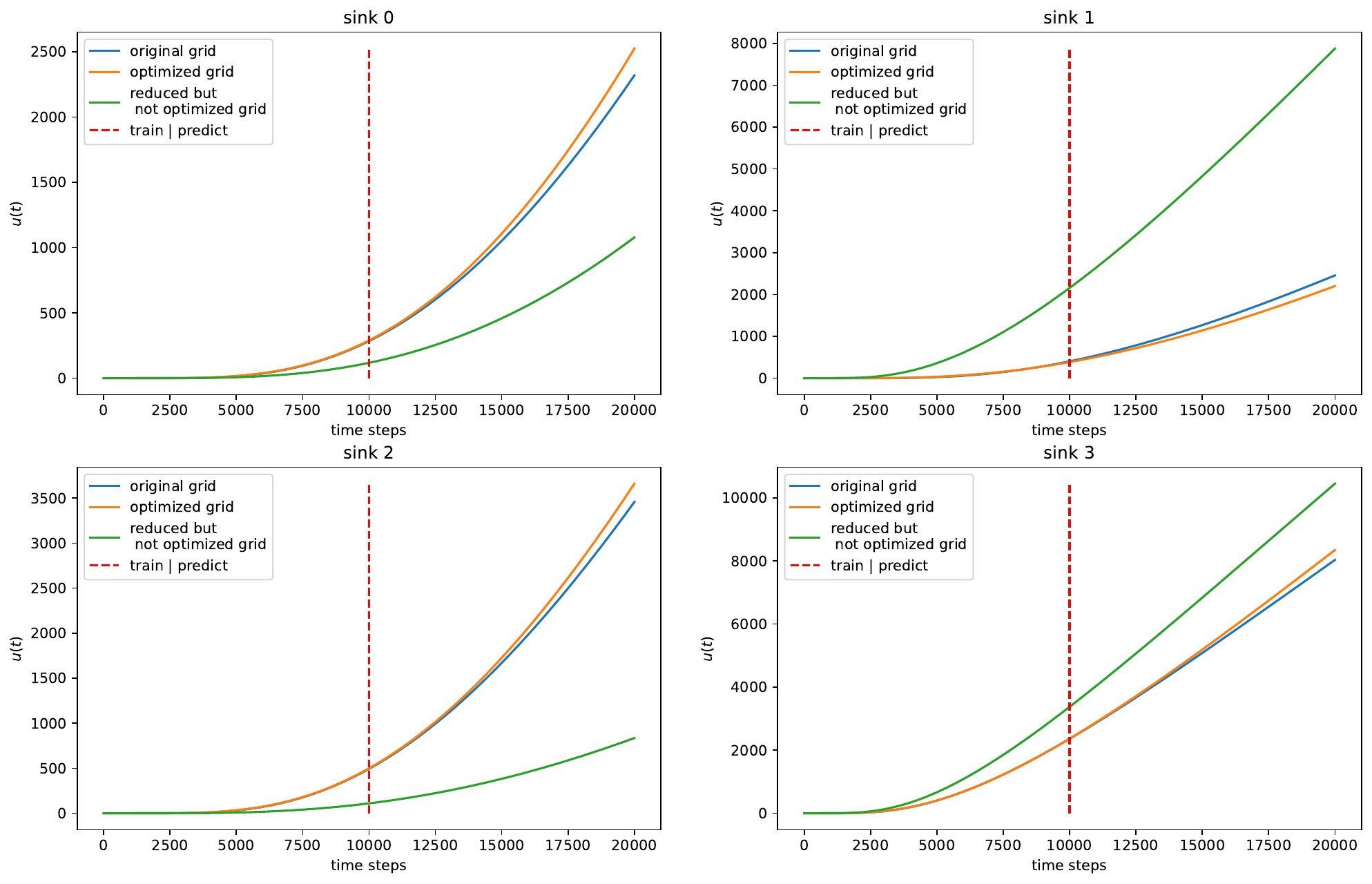}
    \caption{Real-world data experiment with 4 measurement points. We conduct an experiment with 4 measurement points and real-world data permeability field. Each graph represents a measurement point. The locations of measurement point are shown in Fig.~\ref{fig:4sinks_experiment setup}. Red dotted line splits training and test periods. Not optimized grid produces much worse quality of modelling in the measurement points.}
    \label{fig:4sinks_experiment results}
\end{figure}

The results show that for 20000 steps the coarsened model produced accurate predictions compared to the not optimized model (see Fig.~\ref{fig:4sinks_experiment results}).

\end{document}